\documentclass[10pt,a4paper,oneside, titlepage=false]{amsart}
\usepackage{fullpage}

\usepackage{hyperref}
\usepackage[style=numeric,doi=true, url=false,sortcites=true,maxnames=99,giveninits=true]{biblatex}
\usepackage{amsmath,amssymb,amsfonts,amsthm}
\usepackage[ruled, lined, longend]{algorithm2e}
\SetAlgorithmName{Phase}{Phase}{List of Phases}
\usepackage{graphicx}
\usepackage{textcomp}
\usepackage{caption}
\usepackage{subcaption}
\usepackage{tikz}
\usepackage{float}
\usetikzlibrary{arrows}

\newcommand{\cL}{\mathcal{L}}
\newcommand{\cN}{\mathcal{N}}
\newcommand{\cD}{\mathcal{D}}

\newcommand{\RR}{\mathbb{R}}
\newcommand{\ZZ}{\mathbb{Z}}
\usepackage{mathtools}
\usepackage{url}

\usepackage{amsaddr}
\usepackage{graphbox}

\tikzset{
  treenode/.style = {align=center, inner sep=0pt, text centered,
    font=\sffamily},
  arn_n/.style = {treenode, circle, white, font=\sffamily\bfseries, draw=black,
    fill=black, text width=1.5em},% arbre rouge noir, noeud noir
  arn_r/.style = {treenode, circle, red, draw=red, 
    text width=1.5em, very thick},% arbre rouge noir, noeud rouge
  arn_x/.style = {treenode, rectangle, draw=black,
    minimum width=0.5em, minimum height=0.5em}% arbre rouge noir, nil
}

\title[An unsupervised learning approach to evaluate questionnaire data]{An unsupervised learning approach to evaluate questionnaire data - what one can learn from violations of measurement invariance}

\author{Max Hahn-Klimroth$^{\ddagger}$, Paul W. Dierkes, Matthias W. Kleespies}
\address{{\tt \{hahnklim@mathematik,dierkes@bio,kleespies@em\}.uni-frankfurt.de}}
\address{Goethe-University Frankfurt, 13 Max-von-Laue St, Frankfurt 60438, Germany}

\thanks{$^\ddagger$MHK is the corresponding author.}

\begin{document}

\maketitle

% 205/250 words
\begin{abstract}

In several branches of the social sciences and humanities, surveys based on standardized questionnaires are a prominent research tool. While there are a variety of ways to analyze the data, some standard procedures have become established. 
When those surveys want to analyze differences in the answer patterns of different groups (e.g., countries, gender, age, ...), these procedures can only be carried out in a meaningful way if there is measurement invariance, i.e., the measured construct has psychometric equivalence across groups.
As recently raised as an open problem by Sauerwein et al. (2021), new evaluation methods that work in the absence of measurement invariance are needed.

This paper promotes an unsupervised learning-based approach to such research data by proposing a procedure that works in three phases: data preparation, clustering of questionnaires, and measuring similarity based on the obtained clustering and the properties of each group.
We generate synthetic data in three data sets, which allows us to compare our approach with the PCA approach under measurement invariance and under violated measurement invariance.
As a main result, we obtain that the approach provides a natural comparison between groups and a natural description of the response patterns of the groups. Moreover, it can be safely applied to a wide variety of data sets, even in the absence of measurement invariance. Finally, this approach allows us to translate (violations of) measurement invariance into a meaningful measure of similarity.
\end{abstract}

\section*{Keywords}
Clustering, Principle Component Analysis, Unsupervised Learning, Methods in Cross-Cultural Studies, Measurement Invariance

\section{Introduction}
\label{sec:introduction}
In several branches of the social sciences and humanities, a prominent research tool is to conduct surveys using standardized questionnaires.
One reason for the prominence of questionnaire-based studies may be that they are inexpensive, relatively easy to administer, and if the responses are standardized, it is easy to compile the data.
Measurement instruments in such questionnaires can either consist of single questions that measure separate variables, such as questions about preferences or daily activities, or they can consist of multiple questions that can be aggregated into a single value or index. 
In the latter case, it is common to say that all the items in the questionnaire measure the same \emph{construct}. 
They are often used when measuring attitudes \cite{Milfont2010}, connection to nature \cite{Mayer2004} or environmental behavior \cite{Kaiser1998}.
A common research question in both cases is whether several groups differ in their preferences, attitudes, or other environmental psychological measures. 
In this way, differences and similarities between age groups\cite{Lieflnder2013}, between genders \cite{Kleespies2020} or across different disciplines \cite{Feucht2023} can be examined.

There are a variety of ways to analyze and evaluate questionnaire data. 
Over the years, some standard procedures have become established in social science research and have been used in countless studies. 
When a large data set consisting of different psychological variables and constructs has been collected, a structurally simplifying procedure, such as a factor analysis or principle component analysis (PCA), is usually carried out to simplify the interpretation of the results \cite{Yong2013, Costello2005}. 
However, these procedures require that the data collected are also suitable for carrying out such an analysis. 
To verify the applicability of these methods, Bartlett's test of sphericity and the Kaiser, Meyer and Olkin criterion (KMO) are usually applied \cite{Kaiser1970, Bartlett1950-vr}. 
The KMO criterion assesses the sampling adequacy for each variable of the model and for the entire model, and Bartlett's test of sphericity tests whether there are correlations between the single items.
A factor analysis or PCA only makes sense if these criteria are met. 
As a general rule, it is assumed that a factor analysis or PCA can only be applied if the Bartlett's test finds significant deviations from the hypothesis of having no correlations and the KMO is above the value of 0.7 \cite{Dziuban1974-ws}.

In these structure-simplifying procedures, similar items are assigned to the same higher-level factors. 
In the analysis, the individual items of the higher-order factors can be summarized by calculating a mean value. 
If there is only a single factor, this is referred to as a unidimensional model. 
In order to confirm the internal consistency (the inter-relatedness between the test items) and validity of the individual factors or components, Cronbach's alpha is often calculated \cite{Cronbach1951-rb}. 
The mean values of the individual factors can then be used to carry out group comparisons using for example hypothesis tests.

When making comparisons between different groups, it is only possible to carry out these procedures in a meaningful way if there is measurement invariance, i.e. the measured construct shows psychometric equivalence between groups \cite{Putnick2016-ty}. 
For example, it is possible that the perception of a measurement instrument differs between different cultural groups and that the factor analysis therefore produces a different factor structure for each group. 
In this case, there is a lack of measurement invariance and a comparison of the different groups is not easily possible. Verifying measurement invariance is a complex and multi-stage process that involves different stages \cite{Knickenberg2019}. Currently, methods are still needed that allow research data to be analyzed despite the lack of measurement invariance, as recently posed as an open problem \cite{Sauerwein2021-gu}. Such methods of analysis could help to carry out cross cultural studies, of which more are needed especially in environmental psychology \cite{Tam2020}.

\paragraph{\textbf{Our contribution}}
In this contribution, we propose an unsupervised learning-based approach towards such research data.
As already described, standard methods require either a similar data structure in all subgroups, or at least comparable pairwise correlations between the individual items across all groups. However, especially when comparing heterogeneous groups, this cannot be guaranteed, so that the application of the standard methods described is not appropriate.
In addition, missing data pose a major challenge, as the standard approach requires missing data to be replaced by the mean of the questionnaire, regardless of any correlations or similarities between different items. As an alternative, missing values are often simply ignored by excluding cases, although this has a negative impact on the sample size.
Finally, from a more statistical point of view, following the standard approach results in dealing with multiple groups comparisons. With an increasing number of groups, the number of pairwise comparisons also increases, which quickly becomes confusing and very difficult to interpret when there are many groups. In this case, applied error corrections can contribute to the result becoming inaccurate.

Our approach will analyze a questionnaire data set in three steps, more precisely, we describe an algorithm that
\begin{enumerate}
    \item prepares the questionnaire data,
    \item clusters the questionnaires according to their \emph{response types},
    \item measures the similarity between groups using the proportion of questionnaires of each response type in the group.
\end{enumerate}
In the data preparation step, the algorithm takes care of missing values in the original data using $k$-nearest neighbor imputation, and prepares the data for the actual clustering step.
The clustering step clusters the individual questionnaires, and the centroids of the clusters will be called \emph{response types}, as they refer to the \emph{typical questionnaire} in each cluster.
Finally, the proportion of each response type per group provides a very natural measure of similarity between groups, and further statistical analyses that might explain group similarities or differences can be applied based on this quantity.

In this paper, we give examples of this method applied to synthetic data and compare the result with the classical methods when they can be applied. 
We also give examples where our approach can be easily applied, but standard methods fail. 
Of course, the unsupervised learning approach itself (the actual clustering) and the imputation approach (nearest neighbor imputation) are well known and extensively studied methods in the data science community. 
However, the main goal of this paper is to combine these methods and to promote this approach for the evaluation of questionnaire data to a wide range of researchers who evaluate questionnaire data in different fields.

\section{Important Definitions and Notation}

\subsection{Studied Datasets}
Below we describe three synthetic datasets that are used throughout the paper. 
The datasets were created using the numpy package in the Python programming language, and for completeness, the generated data are provided in the Supplementary Material.
The first and second datasets are based on questionnaires consisting of seven items, where each item takes an integer value in $[1,5]$, and the third dataset consists of questionnaires with only three such items.

Questionnaires are often used to measure a \emph{construct} using multiple questions.
In this case, the value of each item $x_i$ in a questionnaire can be described as a noisy measurement of some base value $b \in [1,5] \cap \ZZ$. 
In particular, a questionnaire like $(4,4,4,5,4,5,4)$ is expected to be observed, while $(1,3,5,2,4,1,5)$ is very unlikely to occur if the questionnaire is answered truthfully.
The first data set, $\cD_1$, models this case.

As already explained, in the case of $\cD_1$, the standard PCA can easily be applied because the data in each group show a similar,  one-dimensional component structure, and only the distributions of the different base values of the groups might differ. However, in the other data sets, we model the case that some items do not measure the same construct as the others, thus the measurement invariance is violated.
Formally speaking, in those cases, we are given multiple base scores per questionnaire (each corresponding to a construct).
If, in each group, the same items measure the same constructs, PCA can be conducted and will find multiple relevant components. Now, the groups can be compared on each of those components independently.
However, if different items measure different constructs in different groups, the standard approaches can be applied, but it is not clear how the results can be interpreted \cite{Schmitt2008-sq, Van_De_Schoot2015-hi}.
However, our approach is still a valid method in this case. 
Data set $\cD_2$ models a corresponding questionnaire survey.

Finally, it is possible that a questionnaire does not try to measure a specific construct, but each item corresponds to the opinion on a certain topic.
These topics are usually related in studies, but in principle they could be completely unrelated.
The standard approach of a factor analysis, or a PCA, cannot be used to compare groups on such datasets.
However, with the proposed approach, similarity between groups can be measured quite naturally.
The third dataset, $\cD_3$, is presented as an example of this scenario.

\subsubsection{Case 1: measurement invariance is given} 
The first data set $\cD_1$ describes the situation where all items in the questionnaire measure the same construct in each group, but different groups have different perceptions of the items.
This means that measurement invariance is given along the groups.
Four different groups are simulated and each group contains 1,000 questionnaires.
The groups differ in their perceptions of the construct, or more formally, the choice of the base value $b$ differs between the groups.
Group 1 is supposed to contain mostly questionnaires with high item values, while Group 2 typically gives moderately high answers.
Finally, in Group 3, each opinion is roughly equally distributed, and an average person in Group 4 has either a high or a relatively low base value.

Formally, for each group $i$, 1,000 entries $v \in ([1,5] \cap \ZZ)^7$ are first sampled independently from a probability law $\cL_i$, and in a second step the individual entries are perturbed.
To formally describe the laws from which we sample in the first step, we denote by $\delta_x = \delta_{ (x,x,x,x,x,x,x) }$ the Dirac measure on $\underline{x} \in \ZZ^7$, hence
$$ \mathbb{P}(\delta_x( \tau )) = \mathbf{1}\lbrace \underline{x} =  \tau\rbrace, $$
and let $\mathrm{unif(1,5)}$ denote the uniform distribution on $\lbrace 1, 2, 3, 4, 5 \rbrace^7$. Then,
\begin{align*}
    \cL_1 & = 0.25 \cdot \delta_{ 5 } + 0.55 \cdot \delta_{ 4 } + 0.15 \cdot \delta_{ 3 } + 0.05 \cdot \mathrm{unif(1,5)}, \\
    \cL_2 & = 0.25 \cdot \delta_{ 4 } + 0.45 \cdot \delta_{ 3 } + 0.25 \cdot \delta_{ 2 } + 0.05 \cdot \mathrm{unif(1,5)}, \\
    \cL_3 & = 0.19 \cdot \delta_{ 5 } + 0.19 \cdot \delta_{ 4 } + 0.19 \cdot \delta_{ 3 } + 0.19 \cdot \delta_{ 2 } + 0.19 \cdot \delta_{ 1 } + 0.05 \cdot \mathrm{unif(1,5)}, \\
    \cL_4 & = 0.3 \cdot \delta_{5} + 0.175 \cdot \delta_{ 4 } +  0.175 \cdot \delta_{ 2 } + 0.3 \cdot \delta_{ 1 }  + 0.05 \cdot \mathrm{unif(1,5)}.
\end{align*}
In the second step, the value of each element $x$ is perturbed by the following noise function $F$, so that
$$ F(x) = \max \left(1, \min \left( 5, \left\lfloor x + \cN(0, 0.66) \right\rceil \right) \right).$$
This means that an independent Gaussian noise with mean 0 and variance 0.66 is added to the value of each element and the result is rounded to the nearest integer. Also, values above 5 and below 1 are truncated to 5 and 1, respectively.

\subsubsection{Case 2: violations of measurement invariance}
The second data set $\cD_2$ describes the situation where the items that do not measure the same construct are not the same across groups, and thus measurement invariance is violated. This means that, either the standard approaches do not provide an interpretation of the results, or even may not be applied.
However, our approach can still be used to analyze similarities and differences across groups.

To create $\cD_2$, we extend the first data set $\cD_1$ by three additional groups, each consisting of 1,000 independent samples. 
The typical questionnaire in the additional groups, Group 5, Group 6, and Group 7, is described as follows.
In Group 5, the first six items measure the same construct and typically assign a high value to that construct, but item 7 is expected to be answered with a low score.
Group 6 questionnaires are expected to measure the same construct with moderately small baseline scores on items 1,2,3,5,6,7, but item 4 is expected to be atypically high.
Finally, in Group 7, item 4 is expected to be large, item 7 is typically small, and the other items measure the same construct with either large or small base values.

Formally, we describe the probability laws from which we sample as follows. Again, we denote by $\delta_{ v }$ the Dirac measure on $v$ and by $\mathrm{unif(1,5)}$ the uniform distribution on $\lbrace 1, 2, 3, 4, 5 \rbrace$.
Then,
\begin{align*}
    \cL_5 & = 0.24 \cdot \delta_{ (5,5,5,5,5,5,1) } + 0.24 \cdot \delta_{ (5,5,5,5,5,5,2) } \\ 
            & \qquad + 0.24 \cdot \delta_{ (4,4,4,4,4,4,2) } + 0.24 \cdot \delta_{ (4,4,4,4,4,4,1) } + 0.04 \cdot \mathrm{unif(1,5)}, \\
    \cL_6 &= 0.32 \cdot \delta_{ (1,1,1,4,1,1,1) } + 0.32 \cdot \delta_{ (3,3,3,5,3,3,3) } \\ 
            & \qquad + 0.32 \cdot \delta_{ (2,2,2,5,2,2,2) } + 0.04 \cdot \mathrm{unif(1,5)}, \\
    \cL_7 &= 0.12 \cdot \delta_{ (5,5,5, 5, 5,5, 2) } +  0.12 \cdot \delta_{ (5,5,5, 4, 5,5, 1) } + 0.12 \cdot \delta_{ (4,4,4, 4, 4,4, 1) } \\ 
    & \qquad + 0.12 \cdot \delta_{ (4,4,4, 5, 4,4, 2) } + 0.12 \cdot \delta_{ (2,2,2, 4, 2,2, 1) } + 0.12 \cdot \delta_{ (2,2,2, 5, 2,2, 1) } \\
    & \qquad + 0.12 \cdot \delta_{ (1,1,1, 5, 1,1, 1) } + 0.12 \cdot \delta_{ (1,1,1, 4, 1,1, 1) } + 0.04 \cdot \mathrm{unif(1,5)}. \\
 \end{align*}
Having sampled 1,000 elements for each group independently, the same perturbation as before is applied, meaning that each value is perturbed by  
$$ F(x) = \max \left(1, \min \left( 5, \left\lfloor x + \cN(0, 0.66) \right\rceil \right) \right).$$

\subsubsection{Items are unrelated and show differences between groups} 
The last example data set is called $\cD_3$. 
The main idea is that each item asks about a different construct, which means that we expect the items to be weakly correlated.
In this case, we simulate only three items per questionnaire, so that each questionnaire is represented as a point in $[1,5]^3 \cap \ZZ^3$.
The typical response to each item differs by group.
Four groups are modeled: Group 8 is expected to have very different answers to each question, Group 9 is expected to have a high score for item 1, Group 10 is expected to have a low score for this item, and finally Group 11 is expected to have a relatively high score for item 2.

Formally, for Group $i$ we sample 1,000 questionnaires from the corresponding probability law $\cL_i$ and then perturb the individual entries, but the perturbation will be different from the previous cases.
Analogously as before, let $\mathrm{unif(1,5)}$ denote the uniform distribution on $\lbrace 1, 2, 3, 4, 5 \rbrace^3$ and let

\begin{align*}
   & \sigma_1 = (5,3,1), \quad \sigma_2 = (1,3,5), \quad \sigma_3 = (3,3,3), \quad \sigma_4 = (5,1,3), \quad \sigma_5 = (1,5,3), \quad \sigma_6 = (3,5,1).
\end{align*}
Then define
\begin{align*}
    \cL_{8} & = 0.16 \cdot \delta_{ \sigma_1 } + 0.16 \cdot \delta_{ \sigma_2 } + 0.16 \cdot \delta_{ \sigma_3 } + 0.16 \cdot \delta_{ \sigma_4 } + 0.16 \cdot \delta_{ \sigma_5 } + 0.16 \cdot \delta_{ \sigma_6 } + 0.04 \cdot \mathrm{unif(1,5)}, \\
    \cL_{9} & = 0.4 \cdot \delta_{ \sigma_1 } + 0.02 \cdot \delta_{ \sigma_2 } + 0.12 \cdot \delta_{ \sigma_3 } + 0.4 \cdot \delta_{ \sigma_4 } + 0.02 \cdot \delta_{ \sigma_5 } + 0.02 \cdot \delta_{ \sigma_6 } + 0.02 \cdot \mathrm{unif(1,5)}, \\
    \cL_{10} & = 0.06 \cdot \delta_{ \sigma_1 } + 0.4 \cdot \delta_{ \sigma_2 } + 0.04 \cdot \delta_{ \sigma_3 } + 0.02 \cdot \delta_{ \sigma_4 } + 0.4 \cdot \delta_{ \sigma_5 } + 0.06 \cdot \delta_{ \sigma_6 } + 0.02 \cdot \mathrm{unif(1,5)}, \\
    \cL_{11} & = 0.07 \cdot \delta_{ \sigma_1 } + 0.07 \cdot \delta_{ \sigma_2 } + 0.07 \cdot \delta_{ \sigma_3 } + 0.07 \cdot \delta_{ \sigma_4 } + 0.35 \cdot \delta_{ \sigma_5 } + 0.35 \cdot \delta_{ \sigma_6 } + 0.02 \cdot \mathrm{unif(1,5)}.
\end{align*}

After the sample procedure, we perturb each item's value by the noise function $G$, where
$$ G(x) = \max \left(1, \min \left( 5, \left\lfloor x + \cN(0, 1) \right\rceil \right) \right).$$
Compared to $\cD_1$ and $\cD_2$, the perturbation is slightly larger to be able to observe a variety of different questionnaires representing that all items measure different constructs.

\subsection{Ward's clustering method} \label{sec:ward}
The clustering obtained in this paper is due to performing a standard agglomerative clustering with Ward's minimum variance criterion as the objective function \cite{Ward1963}.
When this method is applied to a data set of size $n$, in the first step of the clustering algorithm, all $n$ data points form their own cluster.
Now, in each step, the two clusters whose merging minimizes the total within-cluster distance are merged, and the cluster center of the new cluster is computed as the point minimizing the sum of squares distance to all points in the cluster. More precisely, in each step, the clustering algorithm must find the pair of clusters that leads to the minimum increase in the total within-cluster variance after merging \cite{Ward1963, Cormack1971}.
Formally, we choose those clusters $A$ and $B$ with centers $c_A$ and $c_B$ that minimize $$ \Delta(A,B) = \frac{|A| |B|}{|A| + |B|} || c_A - c_B ||^2.$$
This clustering approach is quite popular because it usually produces compact and comparably sized clusters \cite{Szmrecsanyi2012-nw}.

However, the algorithm requires a stopping criterion. This can either be calculated automatically, e.g. when the increase of the whithin-cluster variance exceeds a certain threshold, or when a certain number of clusters is reached. We follow the second approach, where the user determines the number of clusters obtained by the method. This choice can be guided by clustering indices.

\subsection{Determining the number of clusters}
\label{sec:clustering_index}

A fairly intuitive, however quite recent approach to determine a suitable number of clusters during agglomerative clustering, is based on the so-called \emph{gap statistic} \cite{Tibshirani2001, Mohajer2011}.
It compares the cluster compactness of a given clustering with a null reference distribution of the data, which is data with no (obvious) clustering. 
The number of clusters suggested by the method is the value for which cluster compactness on the original data is significantly smaller than the cluster compactness on the reference data \cite{Tibshirani2001, Mohajer2011}.
Hence, we are looking for a (local) maximum in a scree plot which plots the number of clusters against the gap value.
Intuitively, this corresponds to "unnaturally large gaps" in a corresponding dendrogram.
A dendrogram, which represents a tree, illustrates the arrangement of clusters produced by an agglomerative clustering process.
The leaves of the tree are the individual data points, and whenever two clusters are merged, an edge is used to visualize the merging. 
The corresponding height, the distance from the leaves, is equal to the "distance" of the cluster centroids at that moment.
The clusters are induced by a horizontal line, so the tree is cut into a forest by removing all lines above this line. 
The height of such a natural horizontal line should correspond to the existence of a large level gap in the dendrogram \cite{Tibshirani2001}.

Of course, there are several other indices which can be used to measure the goodness of a clustering, like the Calinski-Harabasz index \cite{Calinski1974, DBLP:journals/tcyb/LiuLXGWW13} or the Silhouette coefficient \cite{Rousseeuw1987}. However, while the gap statistic can be used with any clustering algorithm, the latter indices are known to prefer convex clusters over non-convex clusters, even if a non-convex variant might intuitively reflect the better clustering \cite{DBLP:journals/tcyb/LiuLXGWW13, Rousseeuw1987}, in particular if an underlying community structure is supposed to exist.
Due to this reason, we decided to use the gap statistic in this contribution.

\section{Evaluation algorithm for questionnaire data}
\label{sec:contribution}

In this section, we present the proposed evaluation method in detail.
As previously described, the algorithm used to evaluate the questionnaire data runs in three phases.
The first phase is the data preparation phase and consists of the following steps.

\vspace*{0.3cm}
\begin{algorithm}[H]
 \KwData{Original data matrix $A$}
 \KwResult{Prepared data matrix $D$}
 Fill missing values by $k$-nearest neighbor imputation and call the resulting matrix $A'$ \\
 Balance group sizes by upsampling rows of smaller groups \\
 Perturb each item value independently by a defined noise function
 \caption{Data preparation}
\end{algorithm}
\vspace*{0.3cm}
The first step of the data preparation phase is called data imputation.
More specifically, there may be questionnaires in which some items were not answered.
Such missing data must be imputed (or the questionnaire ignored).
One method of imputing data that takes into account the correlations between items is $k$-nearest neighbor imputation \cite{cover_1967, Troyanskaya2001}.
It samples the $k$ most similar questionnaires given the existing items, and fills in the missing items with the average score of the items in the sample.
More formally, suppose that the questionnaire data is expressed by $v \in A \subset \ZZ^d$ and $M \subset \lbrace 1, \ldots,d \rbrace$ are the indices of the items with no answers. 
Then, we define $W = \lbrace x_{\bar{M}} \in \ZZ^{d - |M|}| x \in A \rbrace$ and find the $k$ closest points $X_k(v) \subset W$ to $v_{\bar{M}}$.
Finally, set $v_\ell = \mathrm{avg}_{x \in X_k(v)}( x_\ell )$ for $\ell \in M$.

The next step, the data balancing step, is required to obtain a meaningful clustering of the questionnaires.
As we assume that the distribution of questionnaires might vary between groups, and the sample sizes between the groups might also vary, we need to make sure that the questionnaires appearing (only) in groups with a comparably small sample size, are not irrelevant during the clustering step.
A standard approach to guarantee this in (supervised) learning tasks, is to balance the data by oversampling \emph{minority} groups and/or downsampling \emph{majority} groups \cite{Chawla_2002, Menardi2012}.
In our case, we propose to oversample groups with smaller sample sizes until all groups contain equally many questionnaires, as the actual evaluation will only be with respect to original data and not the synthetically oversampled data (see Phase 2). 

The last data preparation step is to perturb each item's value slightly with independent additive Gaussian noise with mean zero and standard deviation 0.1.
The main purpose is that the clustering (e.g. the cluster \emph{centers}) becomes much more stable towards adding or removing single data points if the raw data is augmented. 
This is a well-known principle, not only in clustering, but in various machine learning tasks in which the models generalize much better if the training data is augmented by random noise \cite{Ding2007, DBLP:conf/icce-tw/ZhangKK20, DBLP:conf/iclr/BelinkovB18, DBLP:conf/acl/MinMDPL20}. Moreover, the data matrix becomes full rank as there are no duplicate rows anymore (with high probability) which increases the numerical stability of the computation.

\vspace*{0.3cm}
\begin{algorithm}[H]
 \KwData{Prepared data matrix $D$}
 \KwResult{Cluster centers $R$ (\emph{response types}) and the groups' \emph{fingerprints} per group $F$}
 Obtain $\ell$ cluster centers $R$ using the data $D$ by agglomerative clustering with Ward's linkage \\
 For each row of $A$, calculate the closest cluster center of $V$ and assign the cluster to the corresponding questionnaire. \\
 Calculate the fingerprints types per group $F$
 \caption{Questionnaire clustering}
\end{algorithm}
\vspace*{0.3cm}

The second phase, the clustering phase, starts with clustering the oversampled and augmented questionnaires given by $D$.
As a clustering algorithm, we propose to perform a standard agglomerative cluster analysis with Ward's minimum variance method as the objective function (see Section \ref{sec:ward}).
The number of clusters $\ell$ is determined visually using the gap statistic \cite{Tibshirani2001} based on a screeplot as well as a dendrogram.
As explained earlier, the number of clusters is expected to be at a local maximum in the scree plot. 
The main idea is described in Section \ref{sec:clustering_index}.

Once the clusters are obtained, we compute the corresponding cluster centers (geometrically speaking, the centroid of each cluster) and call these points \emph{response types}.
While mathematically the response types are really just cluster centroids, the name should reflect the fact that we expect a \emph{typical questionnaire} in the cluster to follow that response type. 
We call the set of all response types $R$ and fix an arbitrary order. Let the ordered response types be $r_1, \ldots, r_\ell$.

Now we \emph{forget} the augmented data $D$ and return to using the original (but imputed) data matrix $A'$. 
For each group $i$, we compute the group's \emph{fingerprint} $f_i$ as a point in the $\ell-$dimensional standard simplex as follows. 
For each $j \in \lbrace 1, \ldots, \ell\rbrace$, we let $f_i$ be the proportion of questionnaires in group $i$ in the cluster corresponding to response type $j$.

\vspace*{0.3cm}
\begin{algorithm}[H]
 \KwData{Response types $R$ and fingerprints $F$}
 \KwResult{Group similarity}
 Interpretation of the response types $R$ content-wise  \\
 Measure the similarity of the fingerprints $F$ \\
 (optional) use additional data and methods to analyze the groups' fingerprints
 \caption{Measuring group similarity}
\end{algorithm}
\vspace*{0.3cm}

The last phase of the proposed method combines the explorative data-driven approach with the actual content interpretation.
First, the response types can be interpreted as a typical response to a questionnaire in that cluster.
The fingerprints thus reflect the distribution of people following a certain response type in the different groups.
The more similar two fingerprints are between two groups, the more similar people answered the questionnaires, which is a natural measure of similarity between groups.

In Section \ref{sec:clustering_index}, it was already explained that a dendrogram (a tree representation of a clustering algorithm) can be used to determine how many natural or robust clusters exist.
It also yields a very intuitive description of similarity between data points, as those points whose clusters merge \emph{earlier}, are more similar.
Such a notion of similarity is also standard outside of data science, for example in ecology and evolution such dendrograms are known as phylogenetic trees and show the evolutionary relationships among species \cite{Letunic2006}.

As an optional last step, the fingerprints in combination with the response types may be explained by group specific properties.
This step is not related to cluster analysis, nor is it part of the proposed method, but for completeness we present it here.
For example, suppose the groups are different countries, and the response types are easy to interpret: response type 1 might reflect a high interest in conservation, response type 2 might reflect a high interest in conservation in principle, but some parts of preservation are irrelevant to the people, etc.
Thus, fingerprints with a high value in response type 1 reflect countries where the majority of people are highly interested in conservation, and fingerprints with a high value in response type 2 reflect countries where people are also interested in conservation, but certain aspects are irrelevant. These results could be explained by indices that describe countries, such as wealth indices or a country's forest cover. 
A simple but powerful way to test such hypotheses is to measure the rank correlation between the marginal of the fingerprint representing a particular response type and the corresponding index.

\section{Results}
\label{sec:results}
\subsection{Structure of the construct equals in all groups}
\label{sec:results_sameconstruct}

\subsubsection{Factor analysis or PCA with follow-up testing}
In the first phase, the applicability of PCA was assessed for each of the four groups using Bartlett's test and the Kaiser-Meyer-Olkin (KMO) criterion. Given the significance of the Bartlett's test ($p < 0.001$) and a KMO criterion of over 0.700 in all groups, PCA was considered appropriate. The PCA in all groups showed that the 7 items could be combined into one higher-order component according to the Kaiser criterion \cite{Kaiser1998}. The calculation of the Cronbach's alpha for this component showed a high internal consistency and reliability between the items for all four groups $(\alpha > 0.700)$. After determining the component, the mean values of the items were calculated for each data point. To determine differences between the groups, those mean values were compared using a hypothesis test between groups. We applied a Kruskal-Wallis test followed by the Dunn-Bonferroni post-hoc test. The level of significance was adjusted by the Bonferroni correction. The results demonstrated pairwise significant differences between all groups, with the exception of Groups 2-3 ($p = 0.564$) and 3-4 ($p = 1.00$). 

\subsubsection{Our approach}
\begin{figure}[ht]
   \centering
   \textbf{Determining the number of response types on data set} $\mathbf{\cD_1}$ \par\medskip
   \includegraphics[width = 0.49\textwidth, height = 5cm]{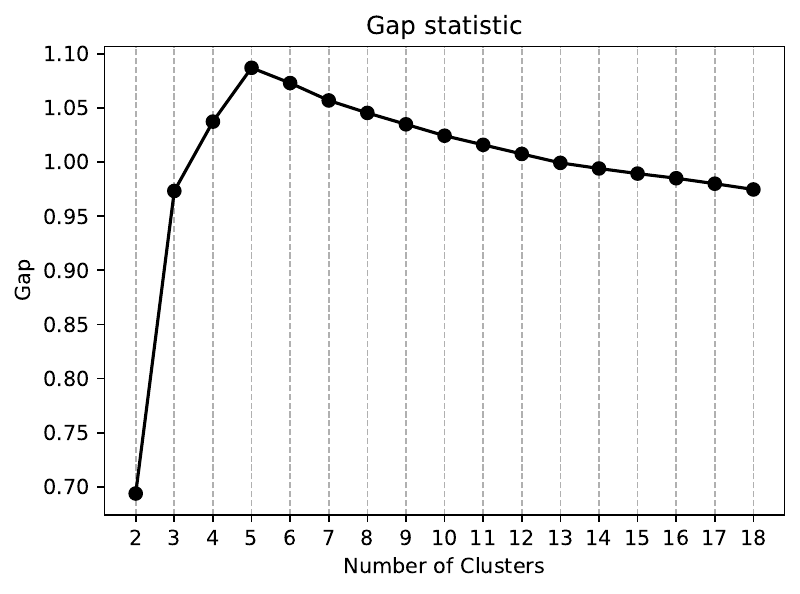}
   \includegraphics[width = 0.49\textwidth, height = 5cm]{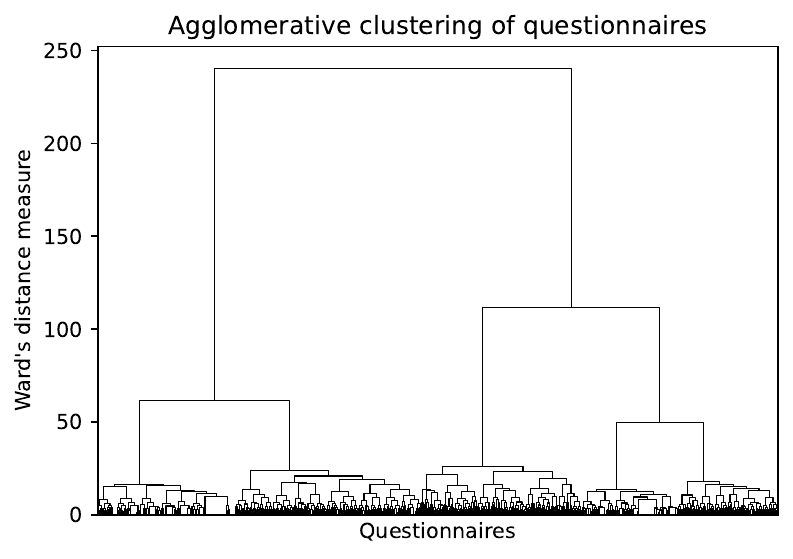} \\
    \textbf{Response types on data set} $\mathbf{\cD_1}$ \par\medskip
   \includegraphics[width = 0.19 \textwidth]{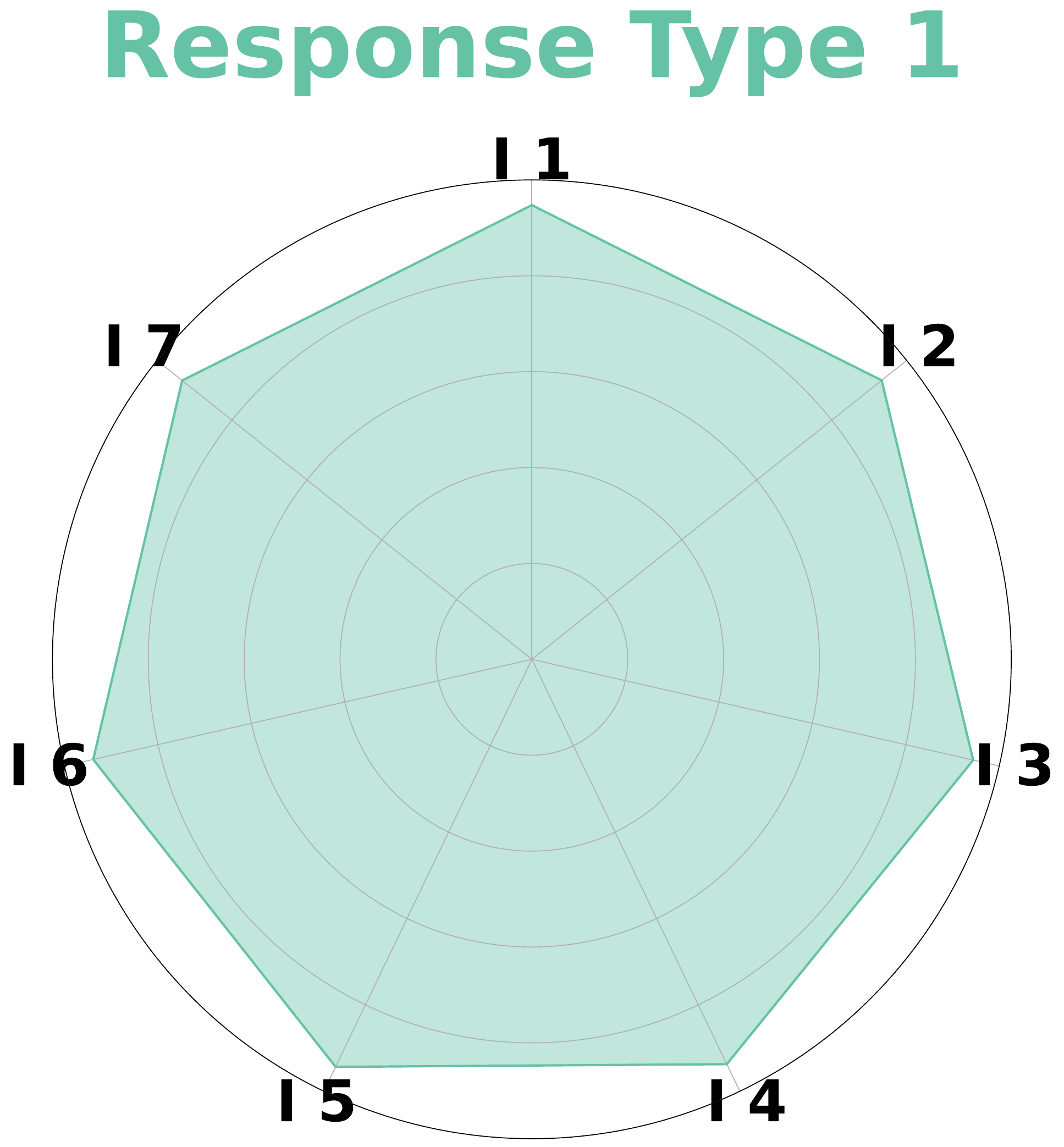} 
   \includegraphics[width = 0.19 \textwidth]{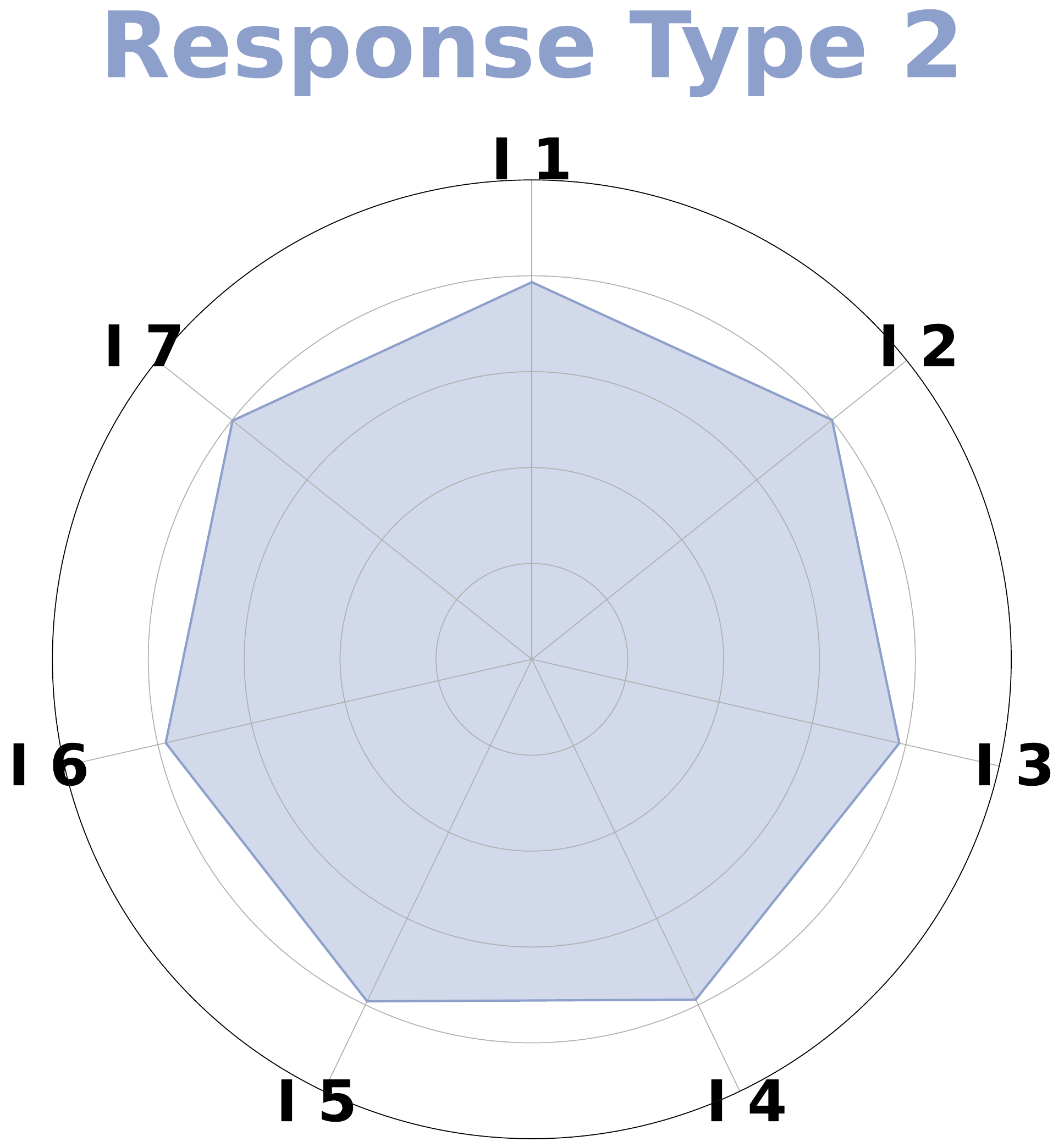} 
   \includegraphics[width = 0.19 \textwidth]{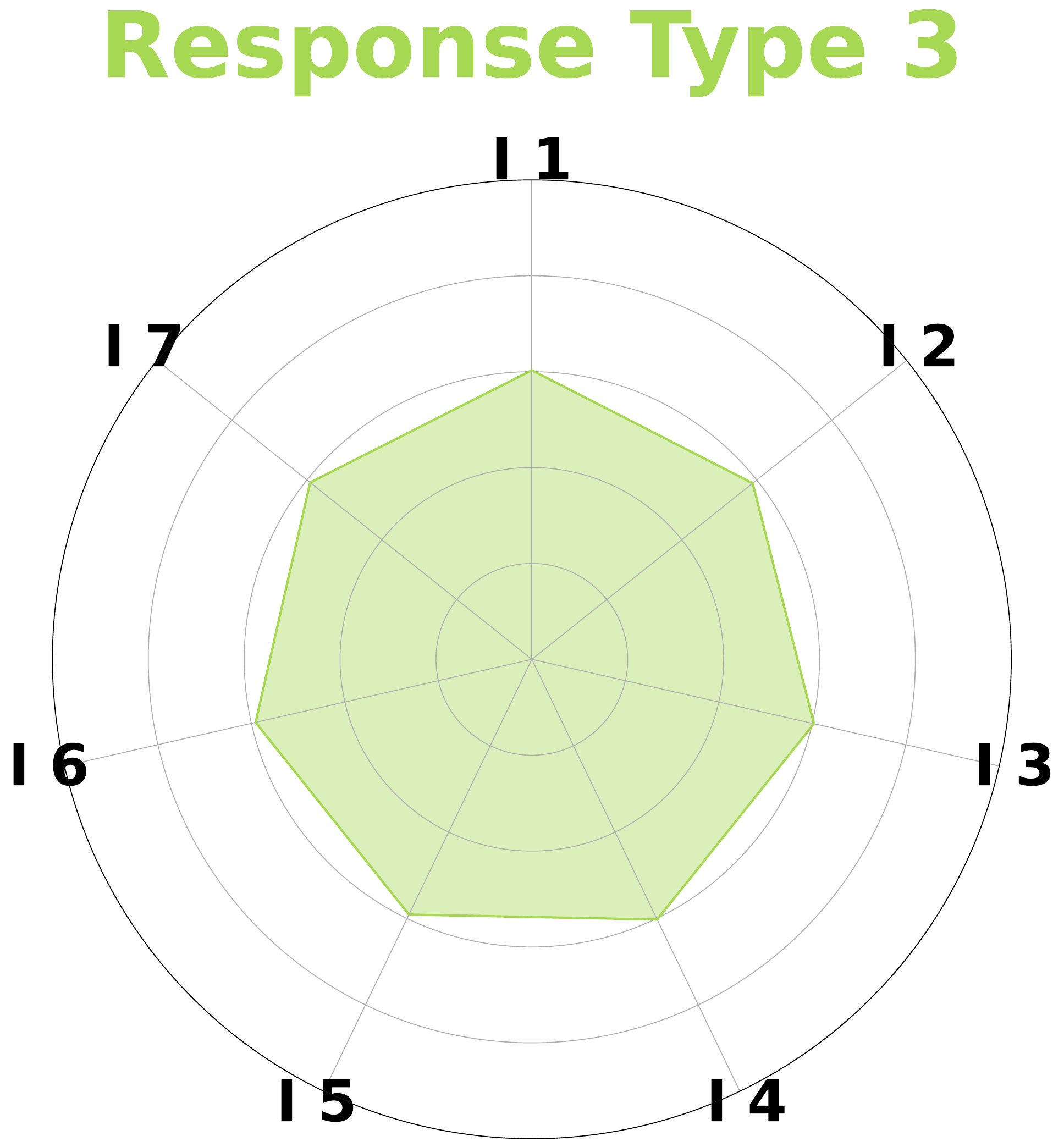} 
   \includegraphics[width = 0.19 \textwidth]{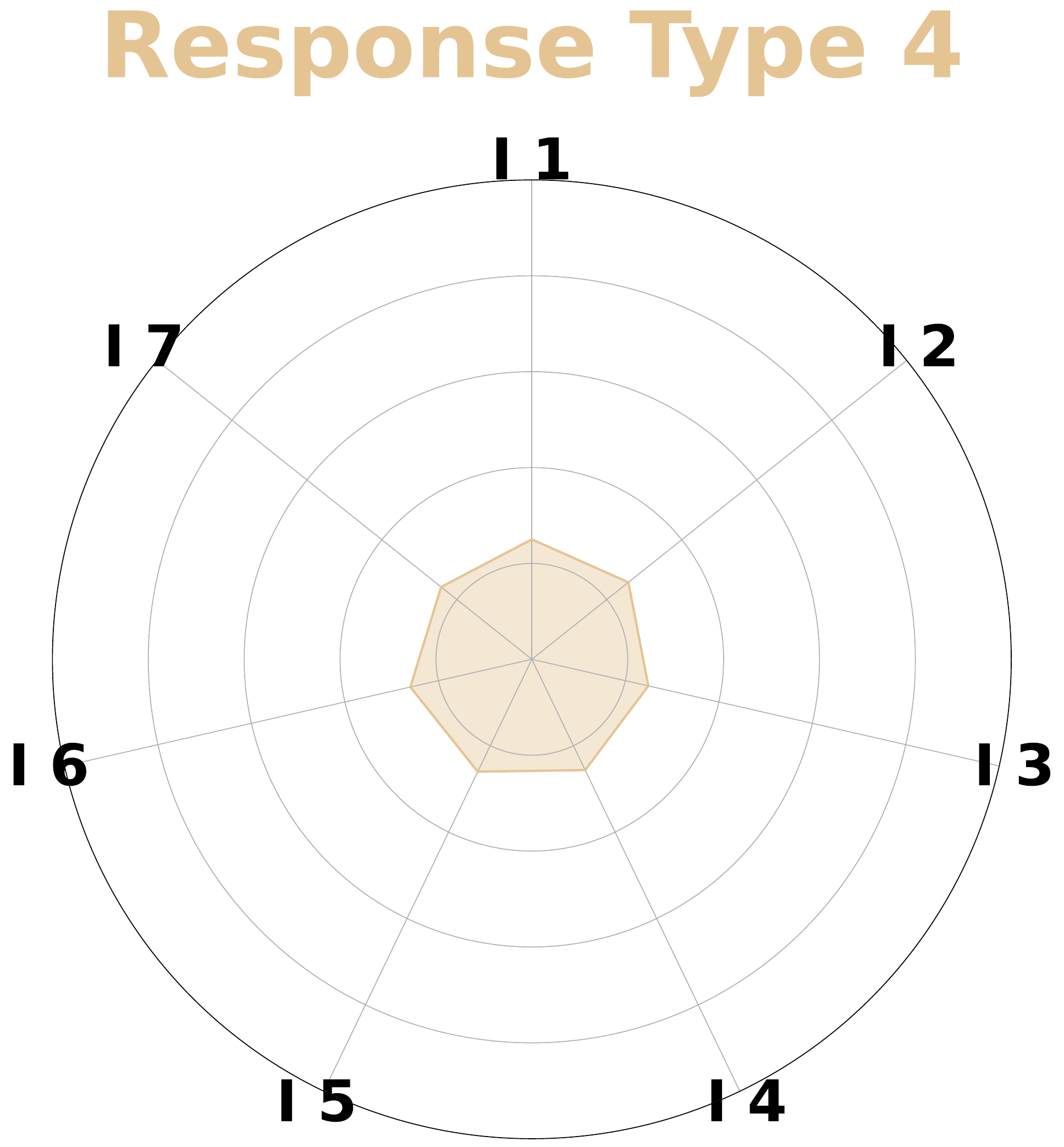} 
   \includegraphics[width = 0.19 \textwidth]{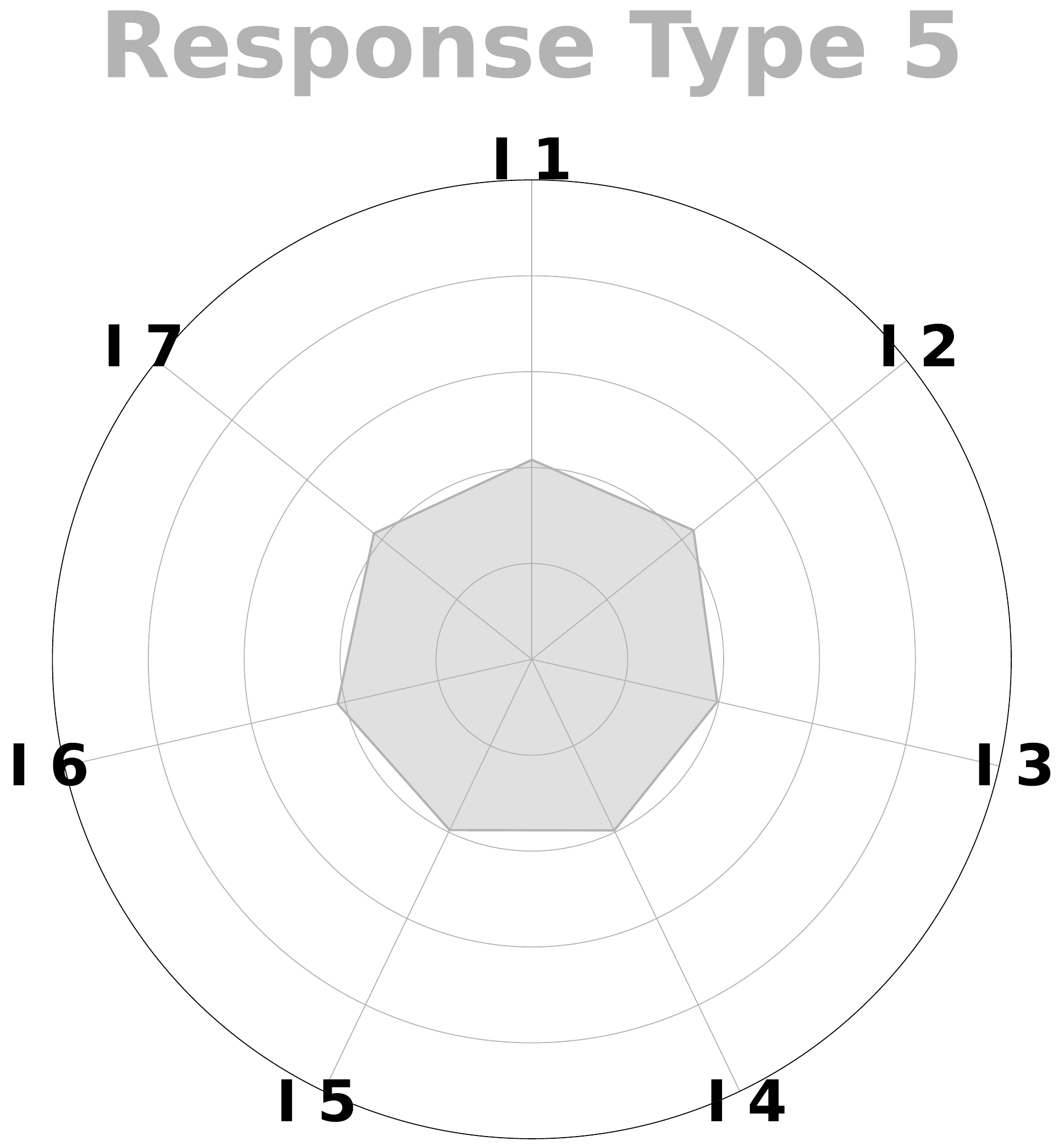} 
   \caption{Graphical representation of the gap statistic as well as the dendrogram corresponding to the goodness of the clustering of the questionnaires in data set $\cD_1$. Moreover, the corresponding response types are shown as a spider plot.}
   \label{fig:clusterindices_d1}
\end{figure}

First, the data preparation described in Phase 1 was applied to the data.
Second, we need to determine an appropriate number of response types. 
Figure \ref{fig:clusterindices_d1} contains the gap values of the gap statistic as well as the corresponding dendrogram.
Clearly, this indicates that 5 response types should be used.
Since the underlying data is artificially generated, we know that we should expect about 5 response types in $\cD_1$, as all items in most questionnaires measure the same construct, but the expected value is different. 
However, due to random fluctuations, we could have observed more response types.
For completeness, we give an example of the stability of our approach with respect to more response types in section \ref{sec:more_clusters}
The response types reflect, as one would expect, the questionnaires $(1, \ldots, 1)$ to $(5, \ldots, 5)$ with some slight noise. 

\begin{figure}[ht]
   \centering
   \textbf{Group similarity on data set} $\mathbf{\cD_1}$ \par\medskip
   \includegraphics[align=c, width = 0.19\textwidth]{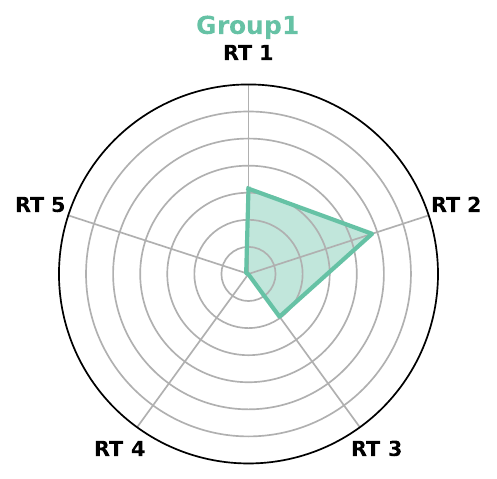}
   \includegraphics[align=c, width = 0.19\textwidth]{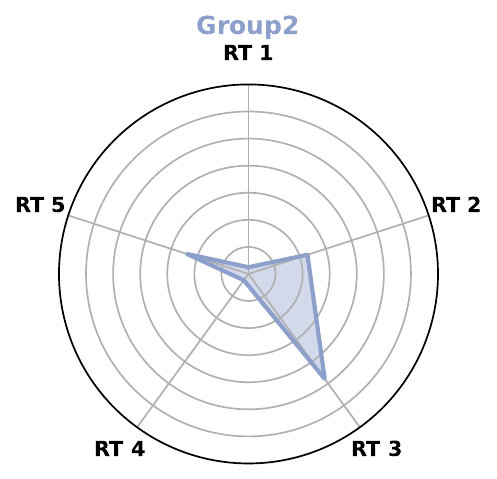}
   \includegraphics[align=c, width = 0.19\textwidth]{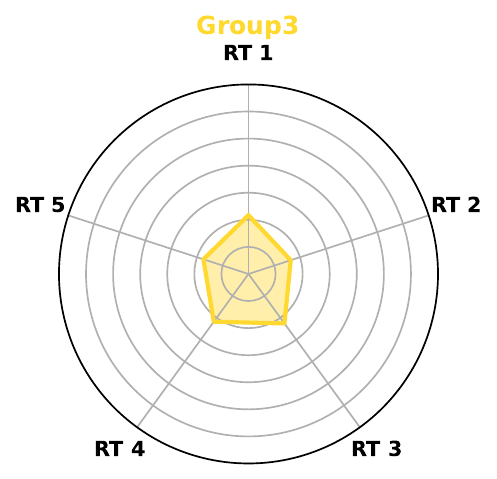}
   \includegraphics[align=c, width = 0.19\textwidth]{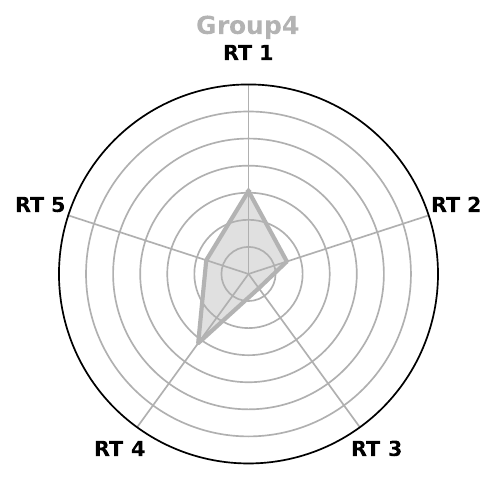}
   \includegraphics[align=c,width = 0.19 \textwidth, height=3cm]{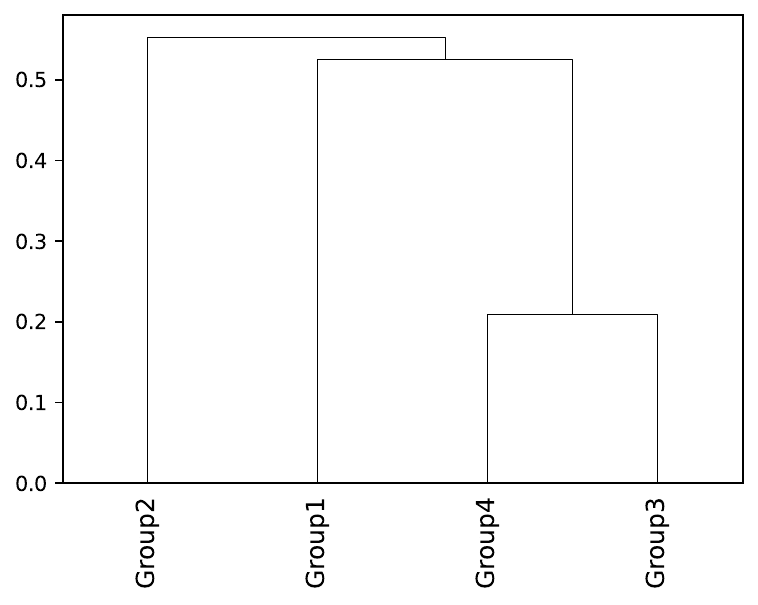}
   \caption{The fingerprints of the different groups regarding the response types as spider plots. The radial $y$-axes are scaled to $(0, 0.7)$. Also, the group similarity on data set $\cD_1$ is given by a dendrogram.}
   \label{fig:similarity_d1}
\end{figure}

The next step is to compute the fingerprints of the 4 groups and express the similarity of the groups (see Figure \ref{fig:similarity_d1}).
The four groups are visually different. Group 1 is concentrated on response types corresponding to large uniform responses, Group 2 is concentrated on medium large answer patterns, Group 3 contains roughly equally many questionnaires of any response type while in Group 4, most questionnaires contain either quite small or quite large answers.
When compared to the model used to generate the data, it is immediately apparent that this is a very good reconstruction of the actual data which is also easy to interpret content-wise.

\subsection{Structures of the construct differs between groups}
\label{sec:results_sameconstruct_variation}
\subsubsection{Factor analysis or PCA with follow-up testing}
In case of the data set $\cD_2$, the standard approach using factor analysis or PCA with subsequent follow-up testing encounters problems. The results of the Bartlett’s test and KMO criterion indicated that a PCA can be performed for Groups 5, 6 and 7 ($p < 0.001$; KMO $> 0.700$). However, several problems arise from the results of the PCAs. In Group 5, item 7 has a much lower loading compared to the other items, while in Group 6, item 4 has a noticeably lower loading. In Group 7, according to the Kaiser criterion, an additional second component is formed by items 4 and 7. As the results of the PCA differ considerably from one another, it is not possible to simply calculate a mean value to compare the groups, as in the previous data set $\cD_1$. A common approach in such cases would be to delete the items with low factor landings or cross-landings from the data set \cite{Yong2013}. 
However, this is likely to result in information being lost and in some cases is not possible due to the low number of items. Especially if measurement invariance is not given, it is not possible to test the constructs between the groups in a meaningful way \cite{Putnick2016-ty}.

\subsubsection{Our approach}
\begin{figure}[ht]
   \centering
   \textbf{Determining the number of response types on data set} $\mathbf{\cD_2}$ \par\medskip
   \includegraphics[width = 0.49\textwidth, height = 5cm]{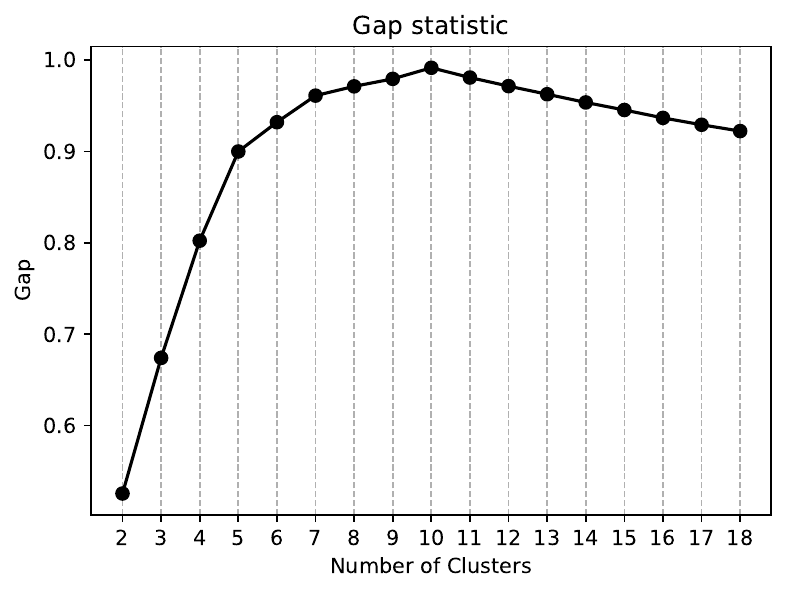}
   \includegraphics[width = 0.49\textwidth, height = 5cm]{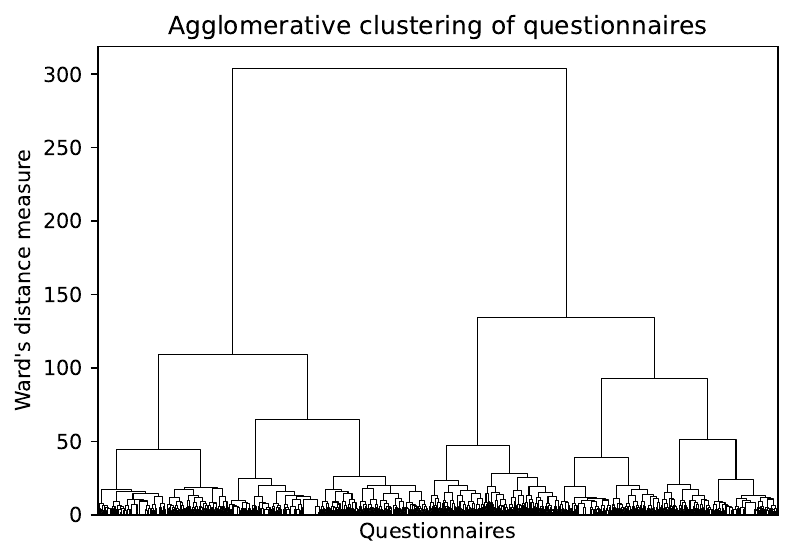} \\
    \textbf{Response types on data set} $\mathbf{\cD_2}$ \par\medskip
   \includegraphics[width = 0.19 \textwidth]{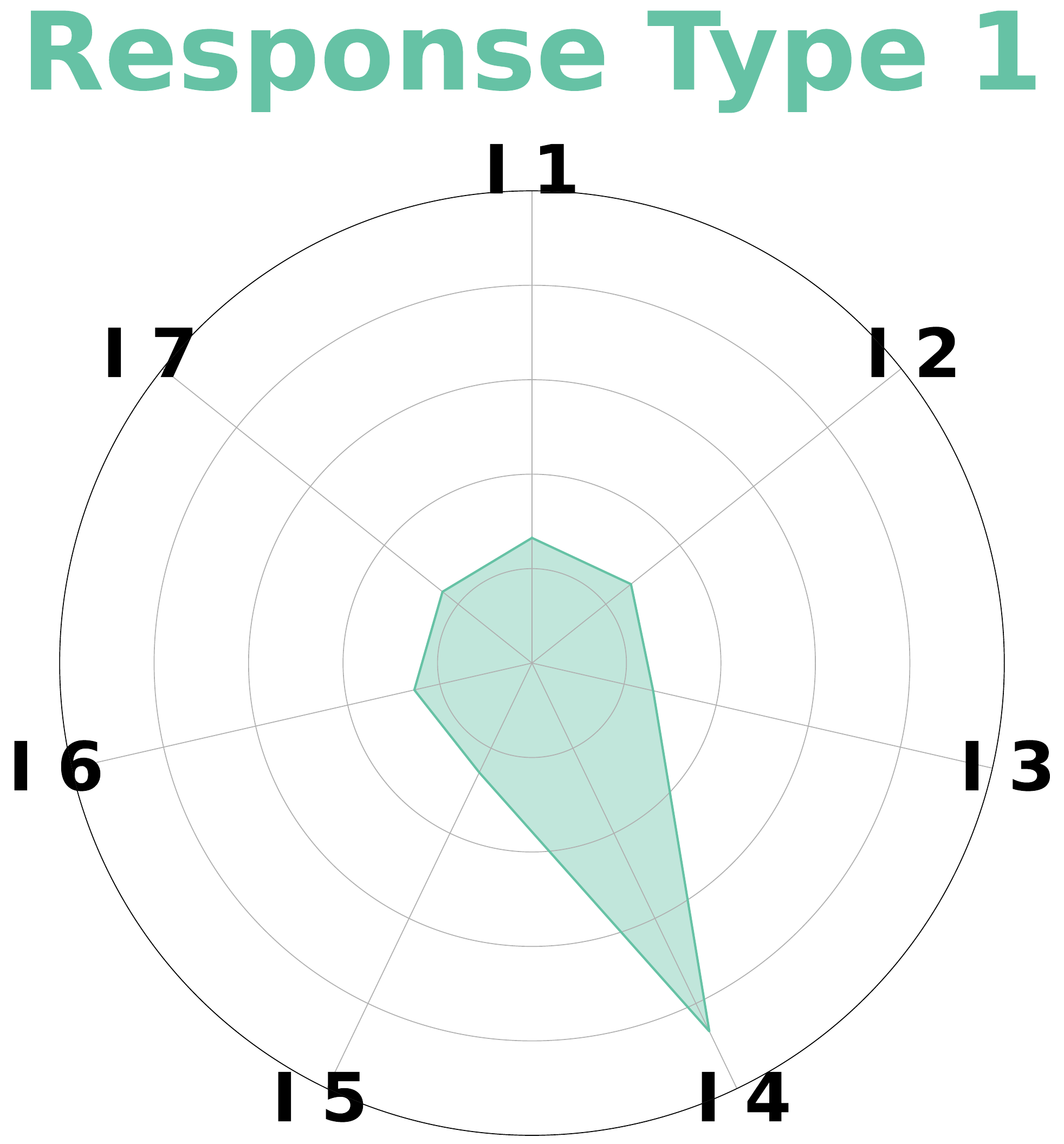} 
   \includegraphics[width = 0.19 \textwidth]{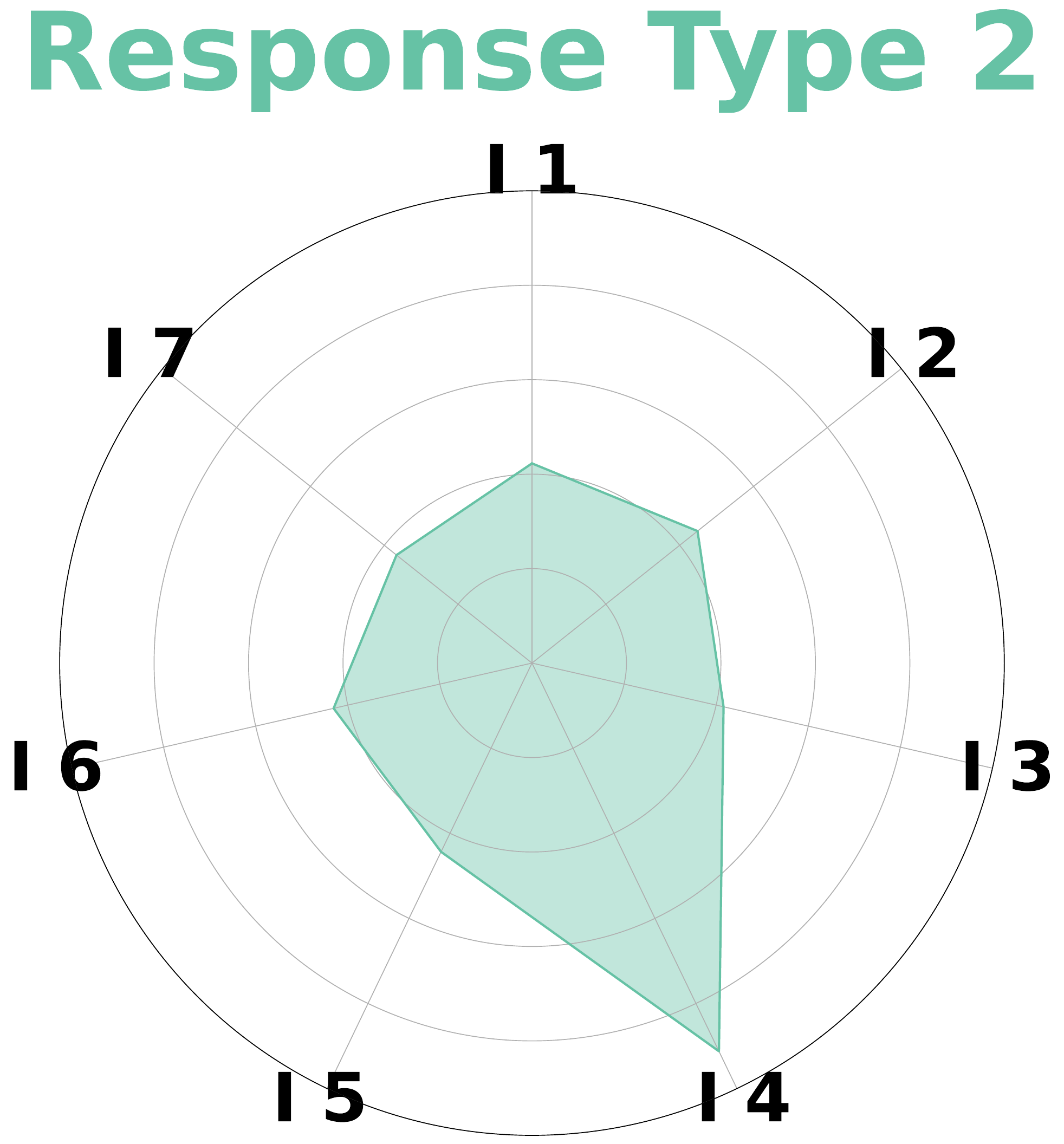} 
   \includegraphics[width = 0.19 \textwidth]{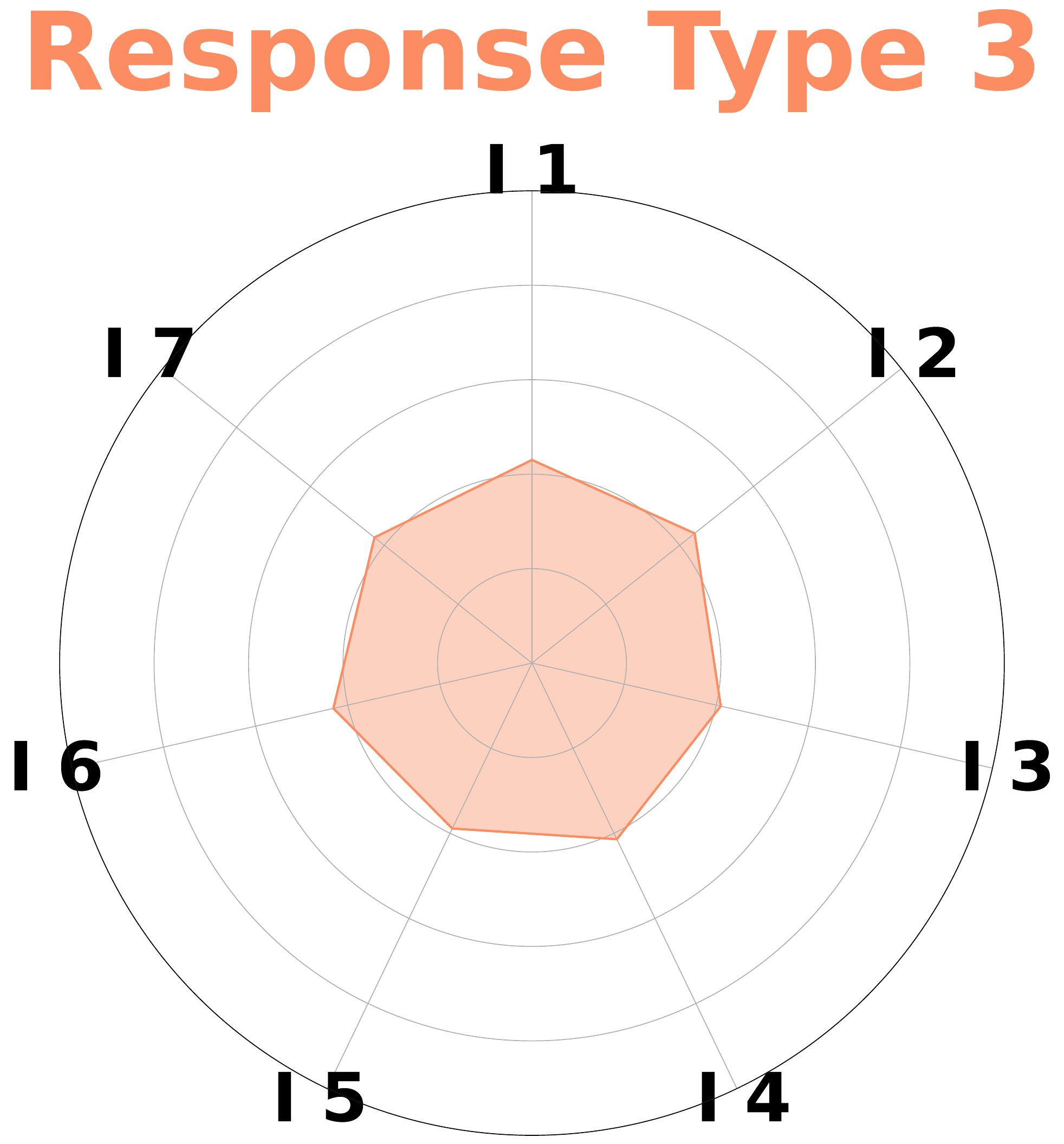} 
   \includegraphics[width = 0.19 \textwidth]{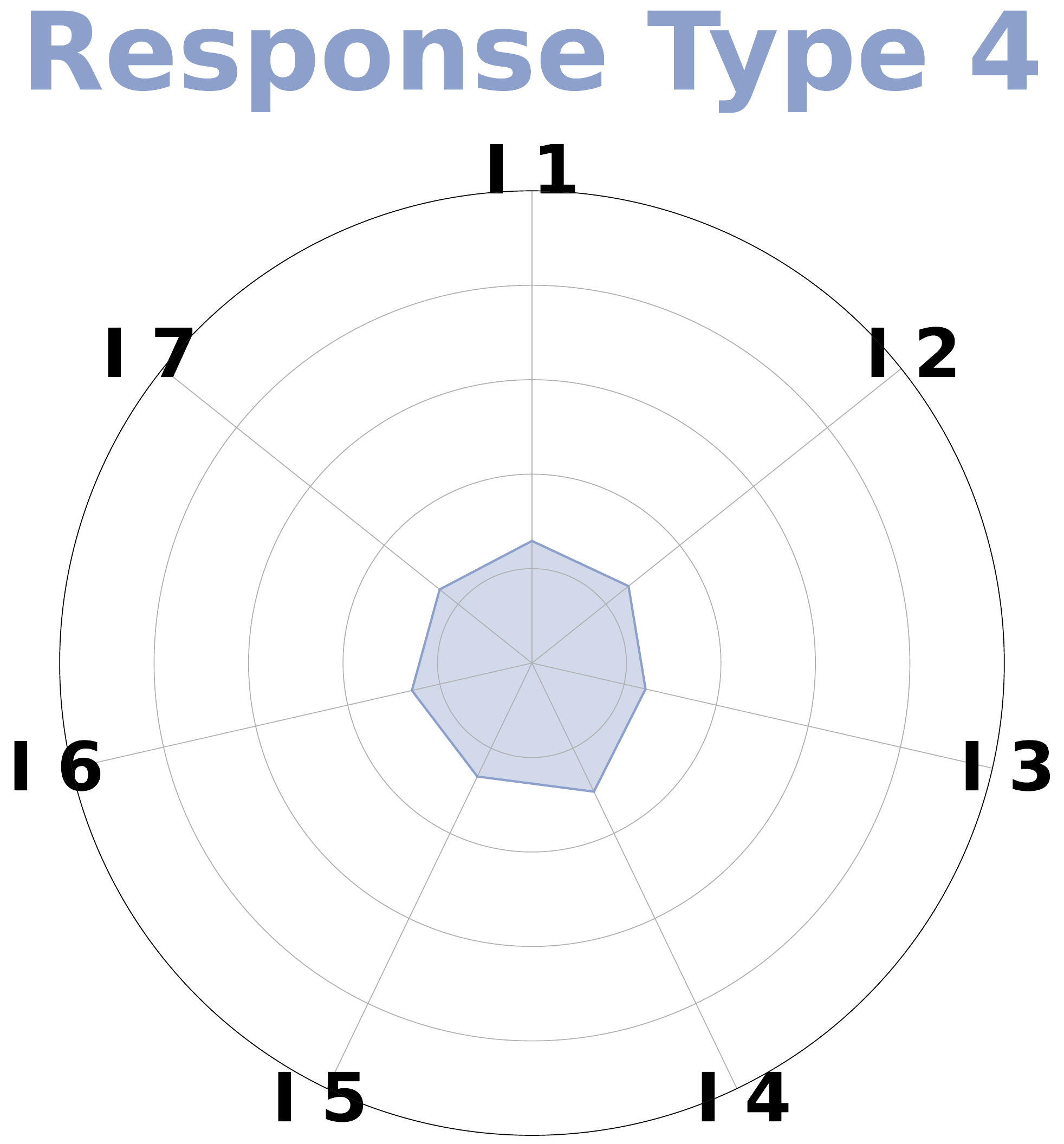}
   \includegraphics[width = 0.19 \textwidth]{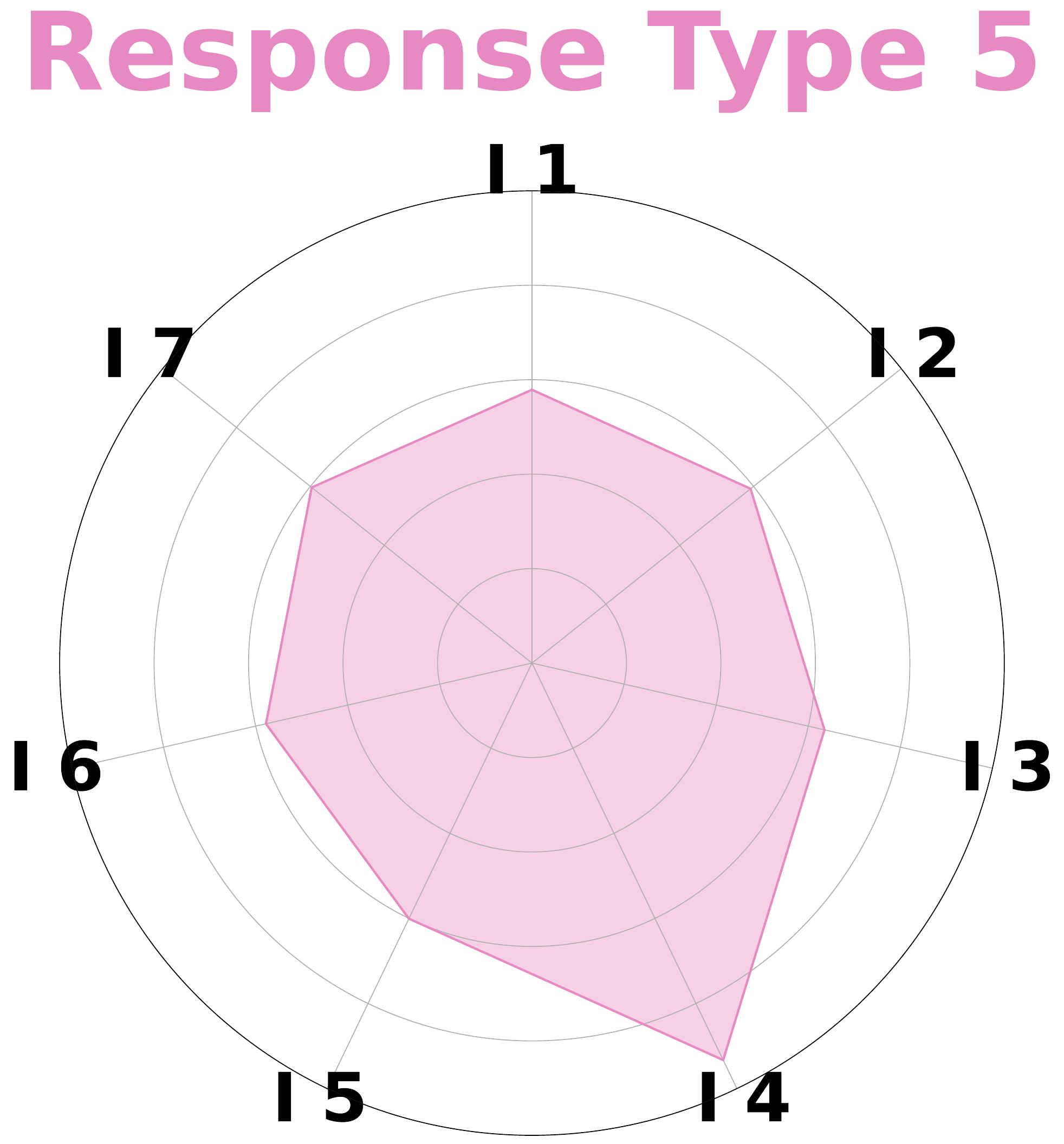} \\
   \vspace*{0.1cm}
   \includegraphics[width = 0.19 \textwidth]{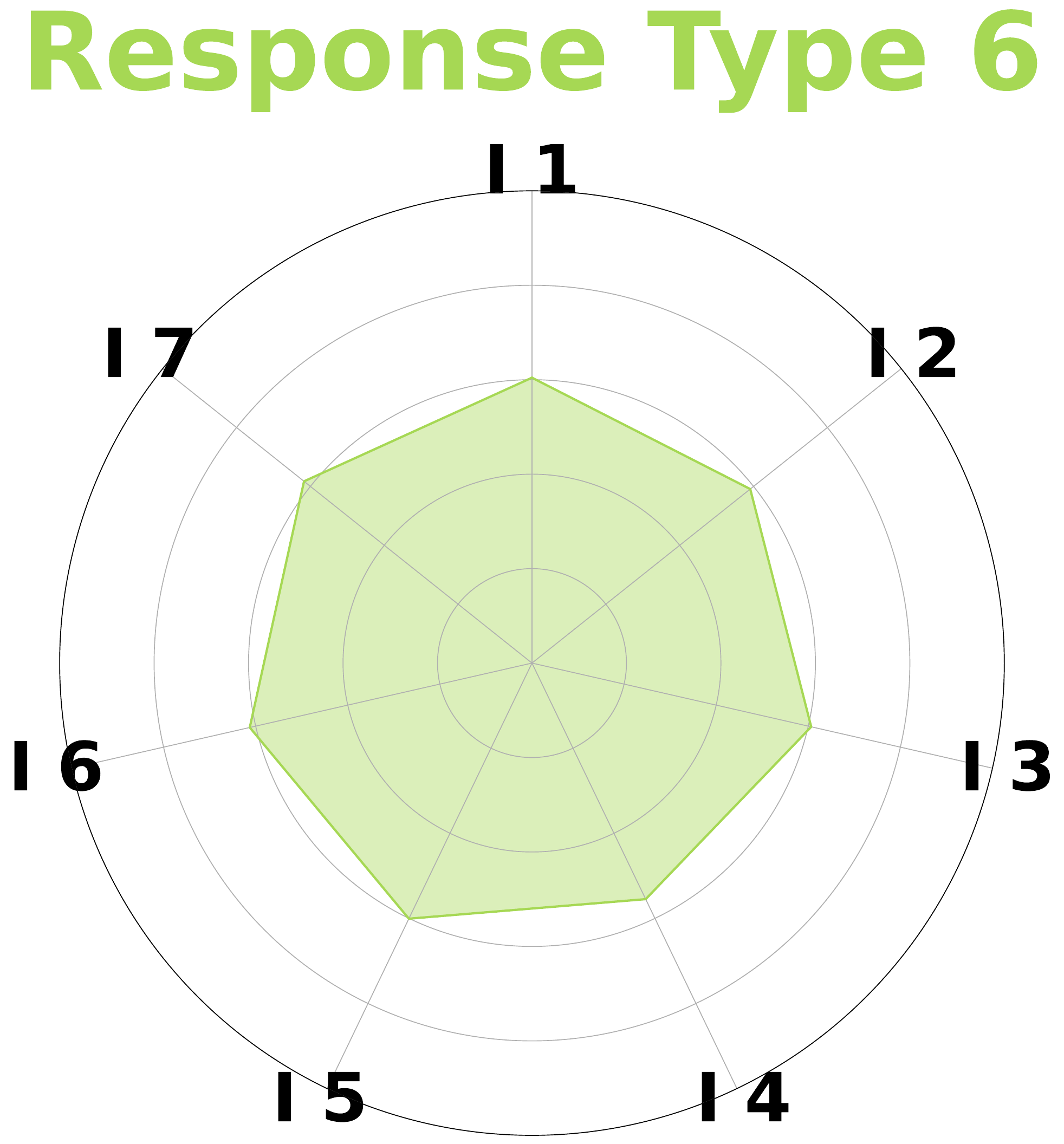} 
   \includegraphics[width = 0.19 \textwidth]{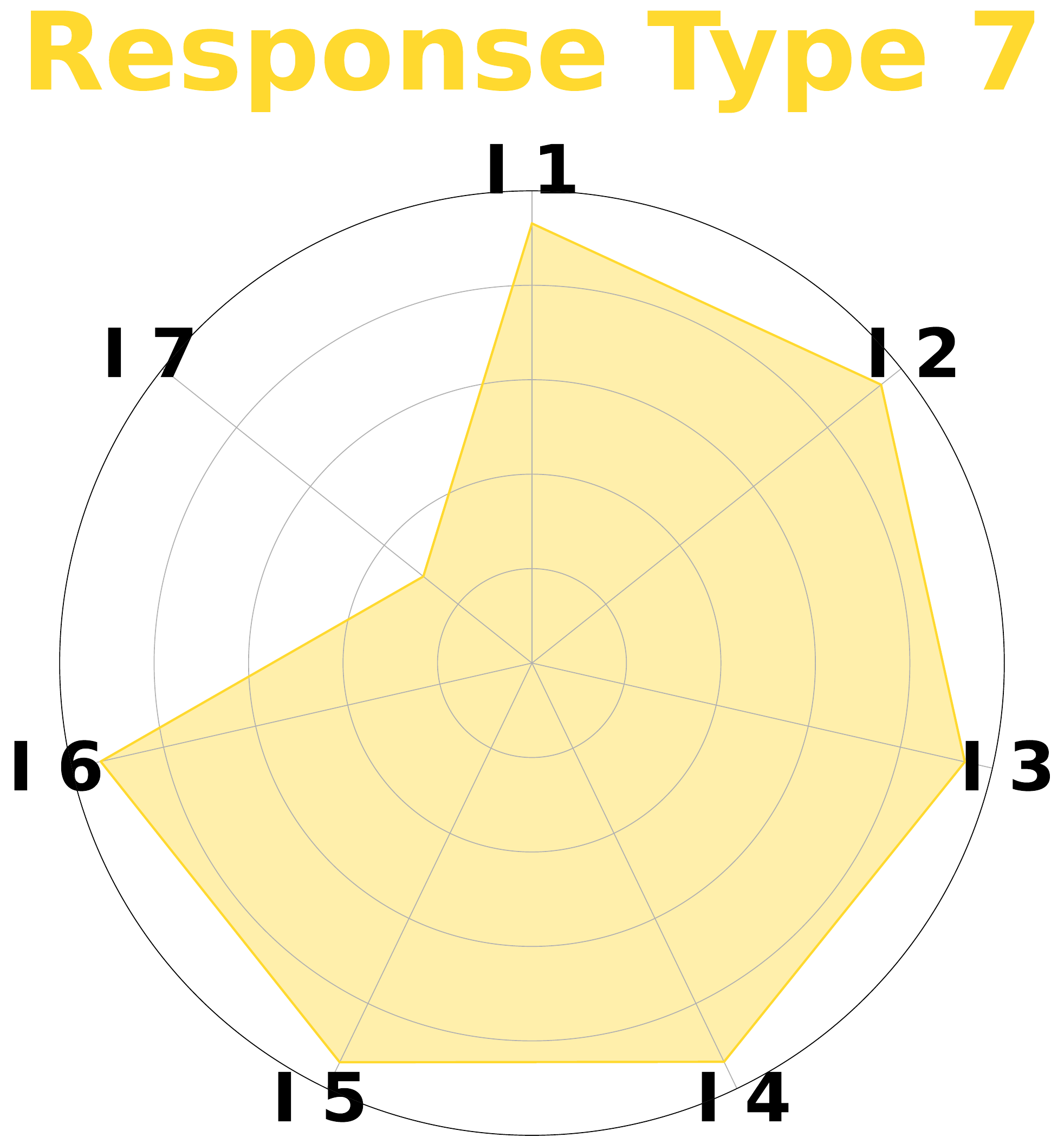} 
   \includegraphics[width = 0.19 \textwidth]{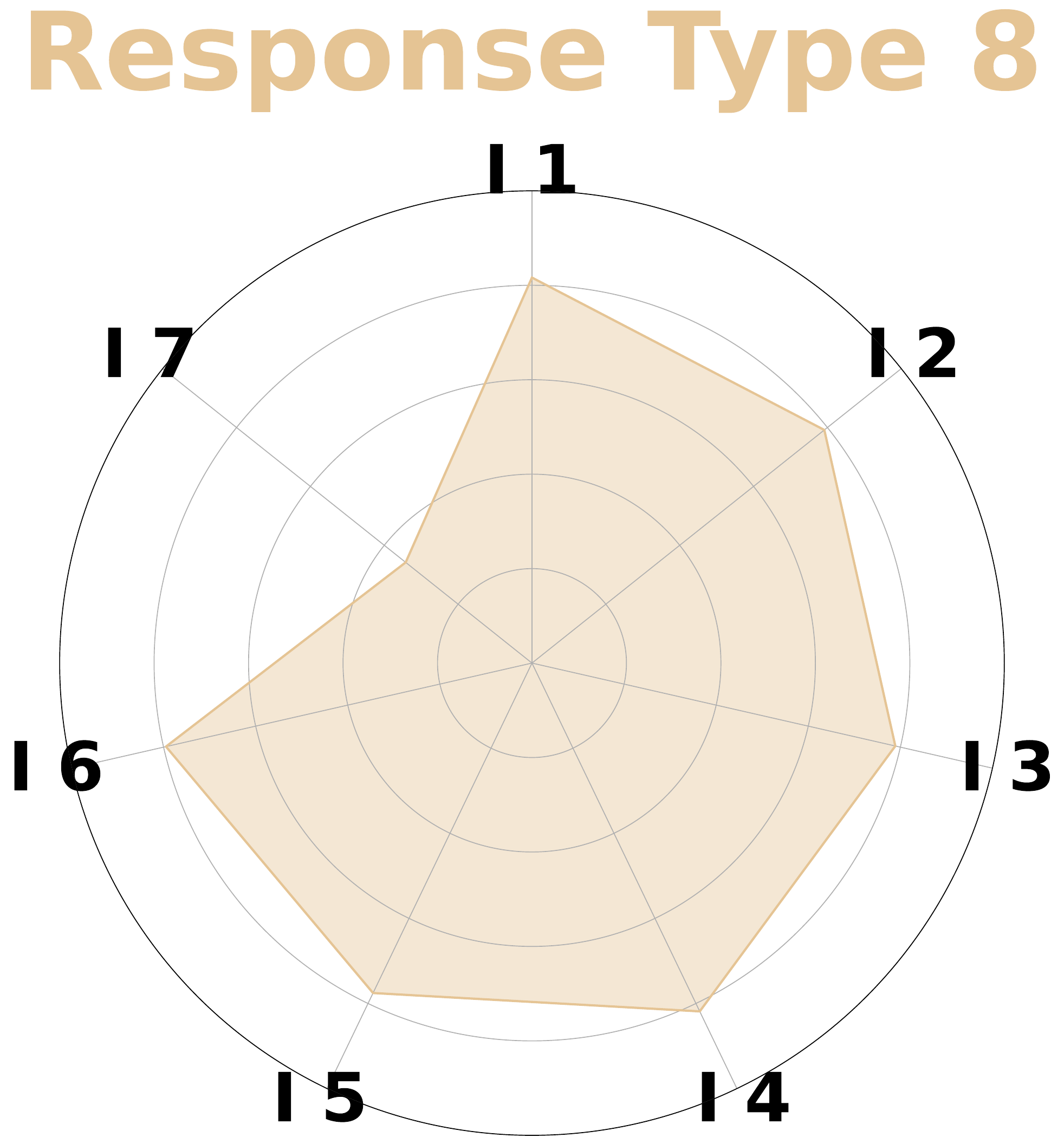} 
   \includegraphics[width = 0.19 \textwidth]{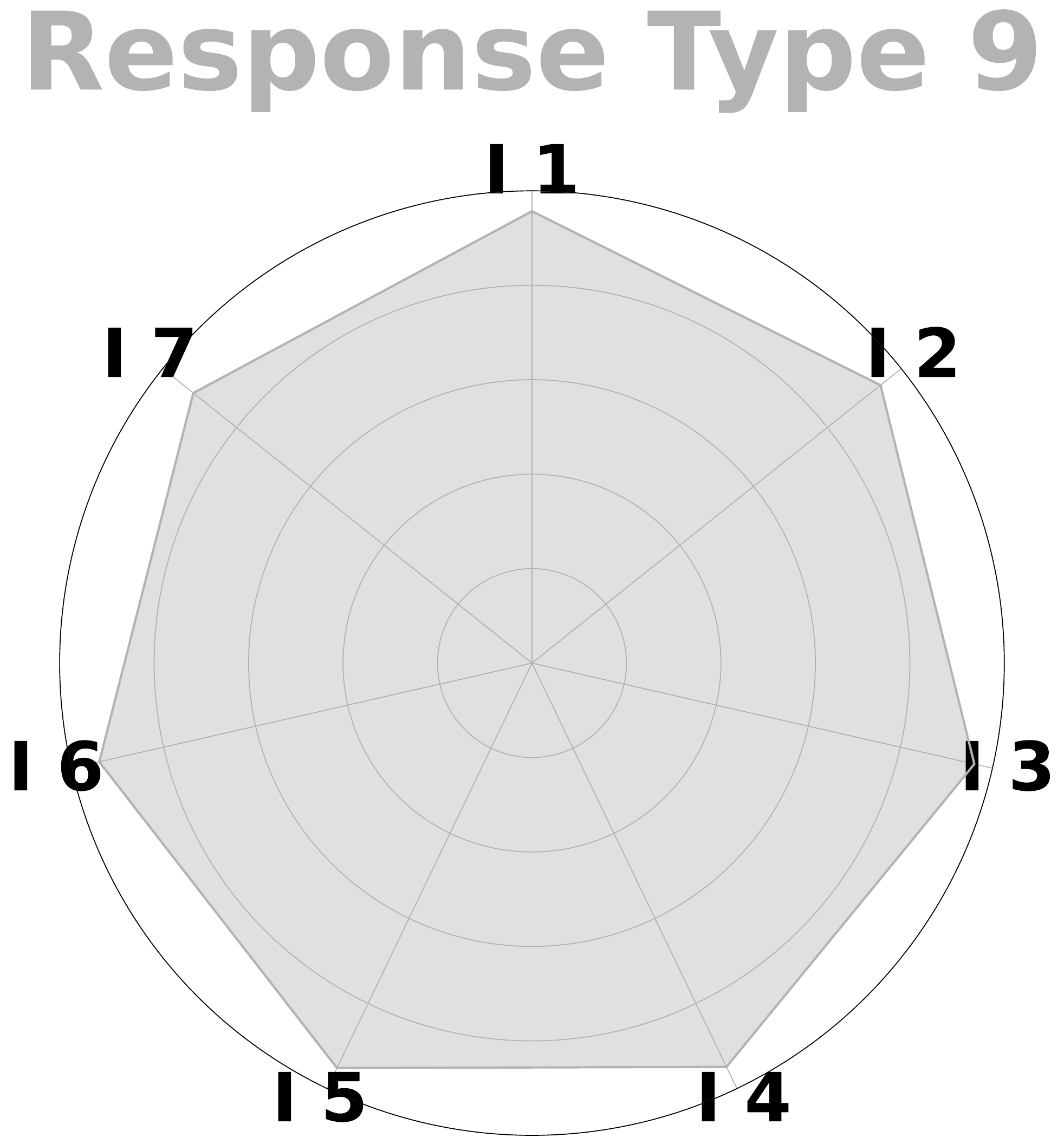} 
   \includegraphics[width = 0.19 \textwidth]{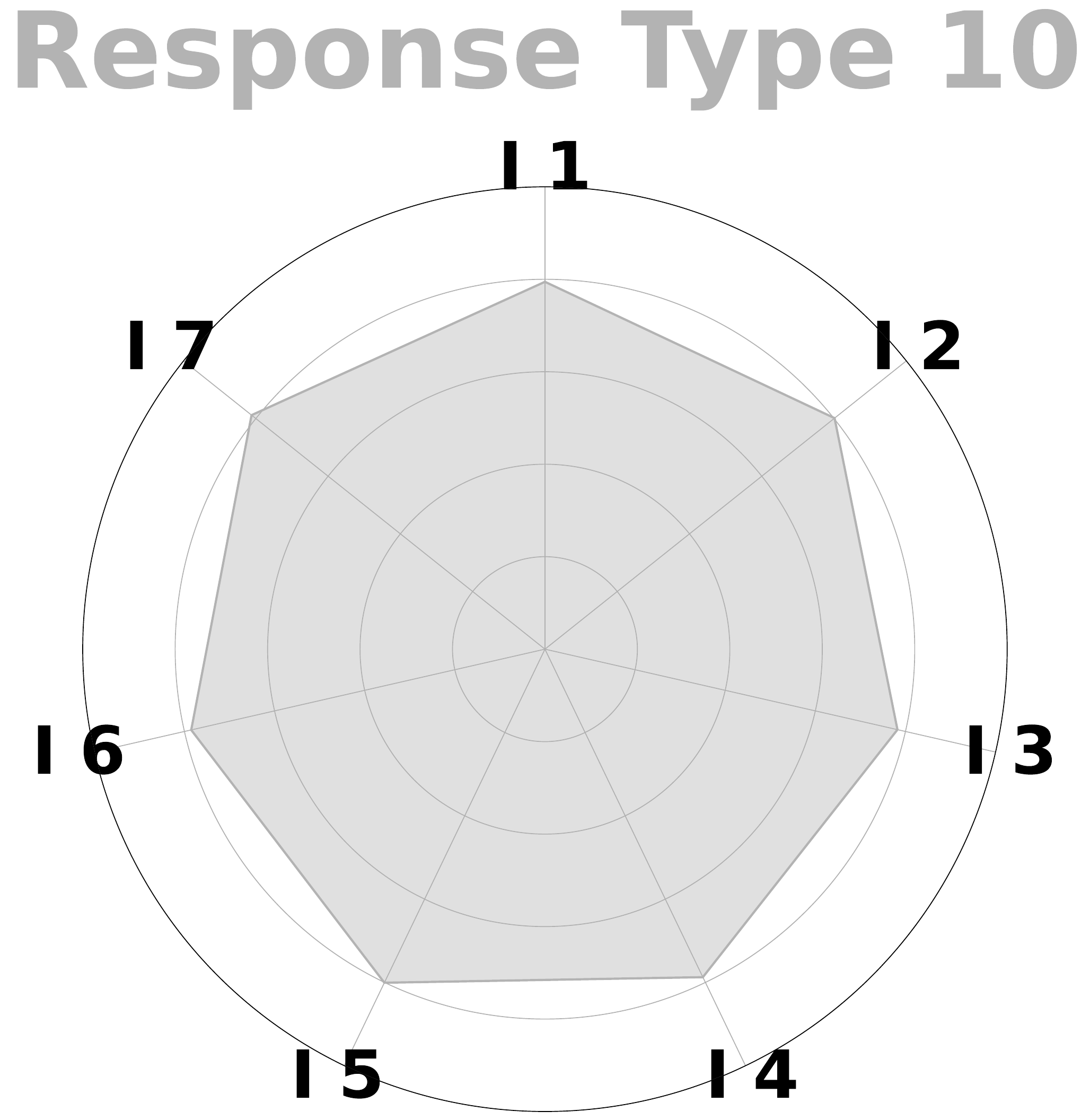}    
   \caption{Graphical representation of the gap statistic as well as the dendrogram corresponding to the goodness of the clustering of the questionnaires in data set $\cD_2$. Moreover, the corresponding response types are shown as a spider plot.}
   \label{fig:clusterindices_d2}
\end{figure}

Again, Figure \ref{fig:clusterindices_d2} gives an overview of the indices that determine the number of response types as well as the similarity of the groups.
The local maximum in the gap statistic is at 10 clusters, and those 10 clusters are highly visible in the corresponding dendrogram.
As in the previous case, this fits well with the model used to generate questionnaires.
We expect up to five \emph{symmetric} clusters in which all items have roughly the same value, as well as clusters in which the typical item is small to moderate but item 4 is large (response types 1, 2 \& 5), and finally clusters in which the typical item is large but item 7 is small (response types 7 \& 8), see Figure \ref{fig:clusterindices_d2}.

\begin{figure}[ht]
   \centering
   \textbf{Group similarity on data set} $\mathbf{\cD_2}$ \par\medskip
   \includegraphics[align=c, width = 0.24\textwidth]{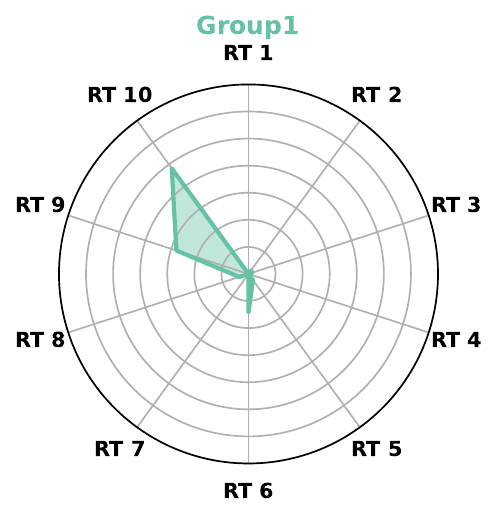}
   \includegraphics[align=c, width = 0.24\textwidth]{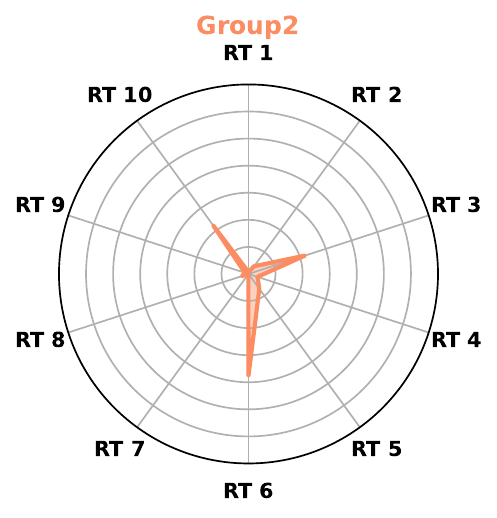}
   \includegraphics[align=c, width = 0.24\textwidth]{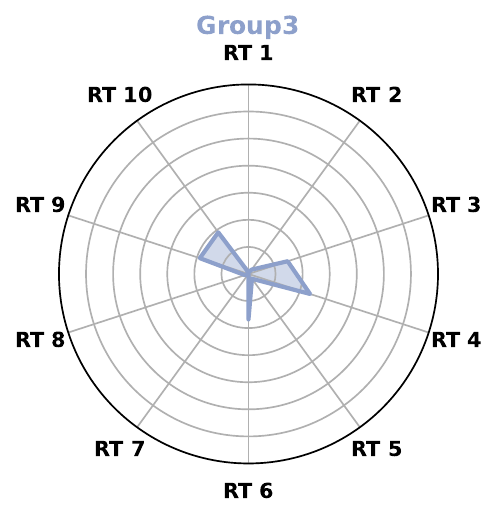}
   \includegraphics[align=c, width = 0.24\textwidth]{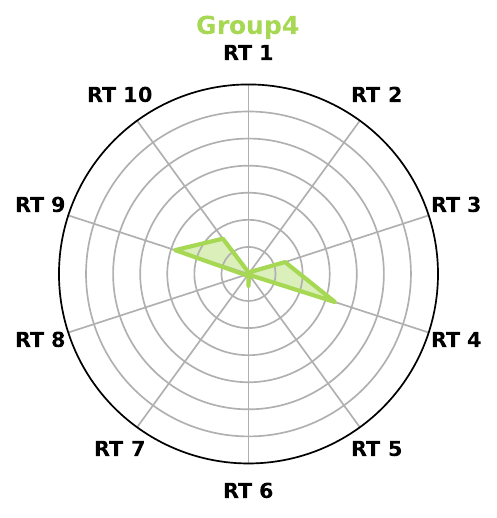}\\
   \includegraphics[align=c, width = 0.24\textwidth]{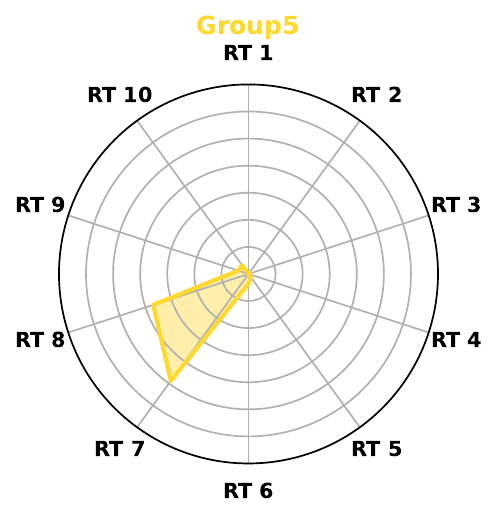}
   \includegraphics[align=c, width = 0.24\textwidth]{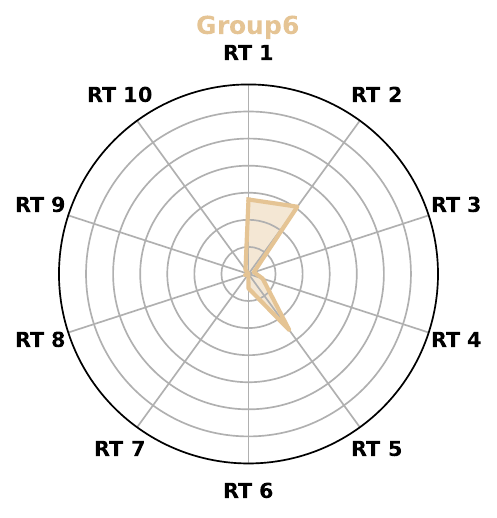}
   \includegraphics[align=c, width = 0.24\textwidth]{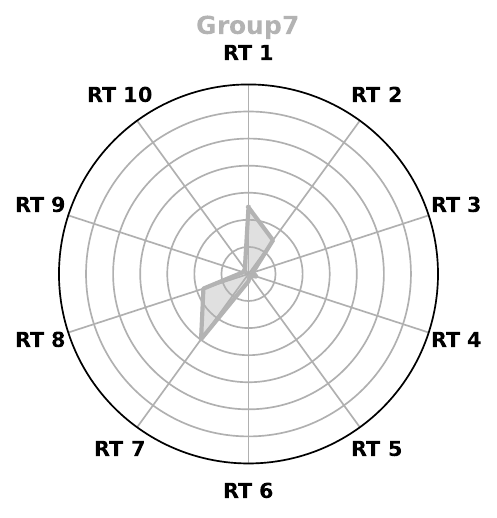}
   \includegraphics[align=c,width = 0.24 \textwidth, height=4cm]{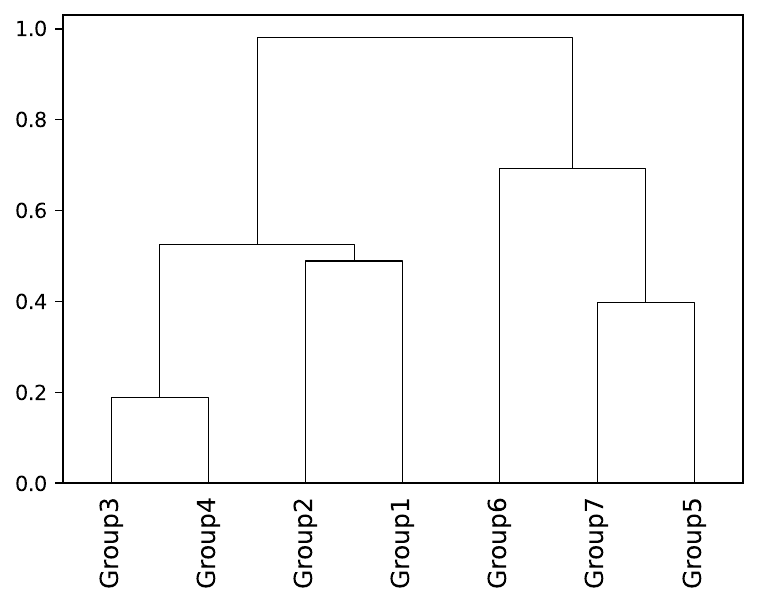}
   \caption{The fingerprints of the different groups regarding the response types as spider plots. The radial $y$-axes are scaled to $(0, 0.7)$. Moreover, the group similarity on data set $\cD_2$ is shown as a dendrogram.}
   \label{fig:similarity_d2}
\end{figure}

Next, the fingerprints of the groups are computed and the similarity of the group is expressed in Figure \ref{fig:similarity_d2}.
Again, the group fingerprints reflect the actual groups very well. Groups 1 to 4 are described similarly to the previous case, they are still concentrated on those response types which express uniform answers of different height. 
Moreover, the "new" groups are also well described, in particular the distribution on the fingerprints yields the following interpretations: Group 5 has many high responses and item 7 is artificially small. Group 6 has small to medium responses but item 4 is large. Finally, in Group 7 we observe a large proportion of questionnaires in which all answers are small but item 4 is large (response types 1 \& 2), but also questionnaires with high answers in which item 7 is comparatively small (response types 7 \& 8). 
We also observe that the previously more similar groups (Group 3 and Group 4) are measured as similar again, and Group 1 and Group 2 are still more similar than either of them is to Group 3 or Group 4. 
The measure of similarity is thus stable with respect to adding data of more groups.

\subsection{Items are unrelated and show differences between groups} 
\label{sec:results_different_constructs}
\subsubsection{Factor analysis or PCA with follow-up testing}
When items show only a low correlation with each other, for example because they measure completely different constructs, a PCA or factor analysis is not applicable. For $\cD_3$, the KMO criterion for all four groups indicates that the conditions for applying a PCA are not met (KMO $< 0.4$). Therefore, the standard approach described above is not applicable for this data set. If a PCA is nevertheless conducted, it produces results that cannot be meaningfully interpreted. The items form two components for each of the groups, which cannot be separated from each other due to high cross-loadings. In addition, the components differ between the four groups. Standard rotation methods (such as varimax) do not improve the result. A meaningful evaluation, a group comparison or a follow-up test is therefore not possible using this method.

\subsubsection{Our approach}
\label{sec:results_diff_constructs} 
\begin{figure}[ht]
   \centering
   \textbf{Determining the number of response types on data set} $\mathbf{\cD_3}$ \par\medskip
   \includegraphics[width = 0.49\textwidth, height = 5cm]{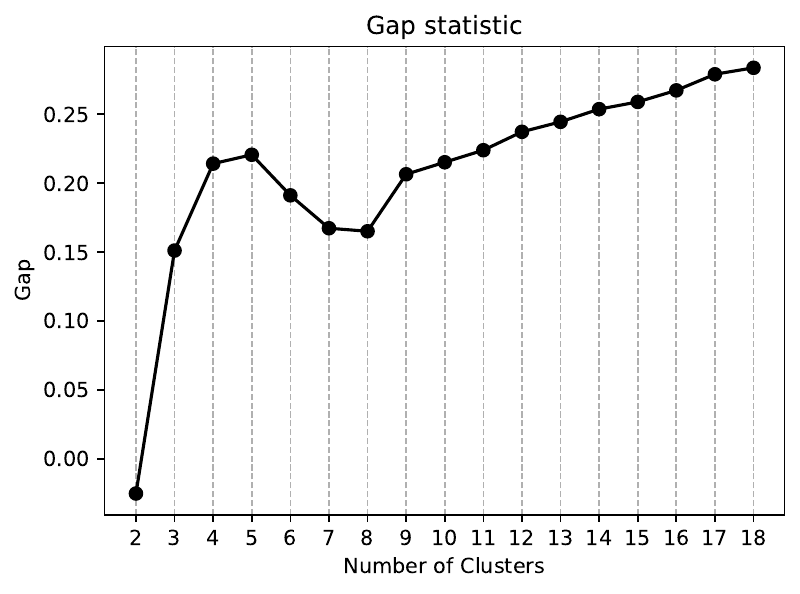}
   \includegraphics[width = 0.49\textwidth, height = 5cm]{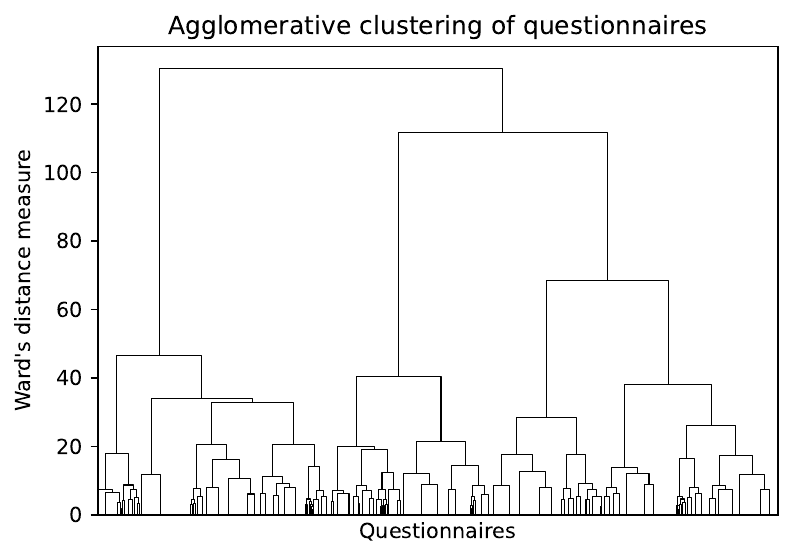} \\
    \textbf{Response types on data set} $\mathbf{\cD_3}$ \par\medskip
   \includegraphics[width = 0.19\textwidth]{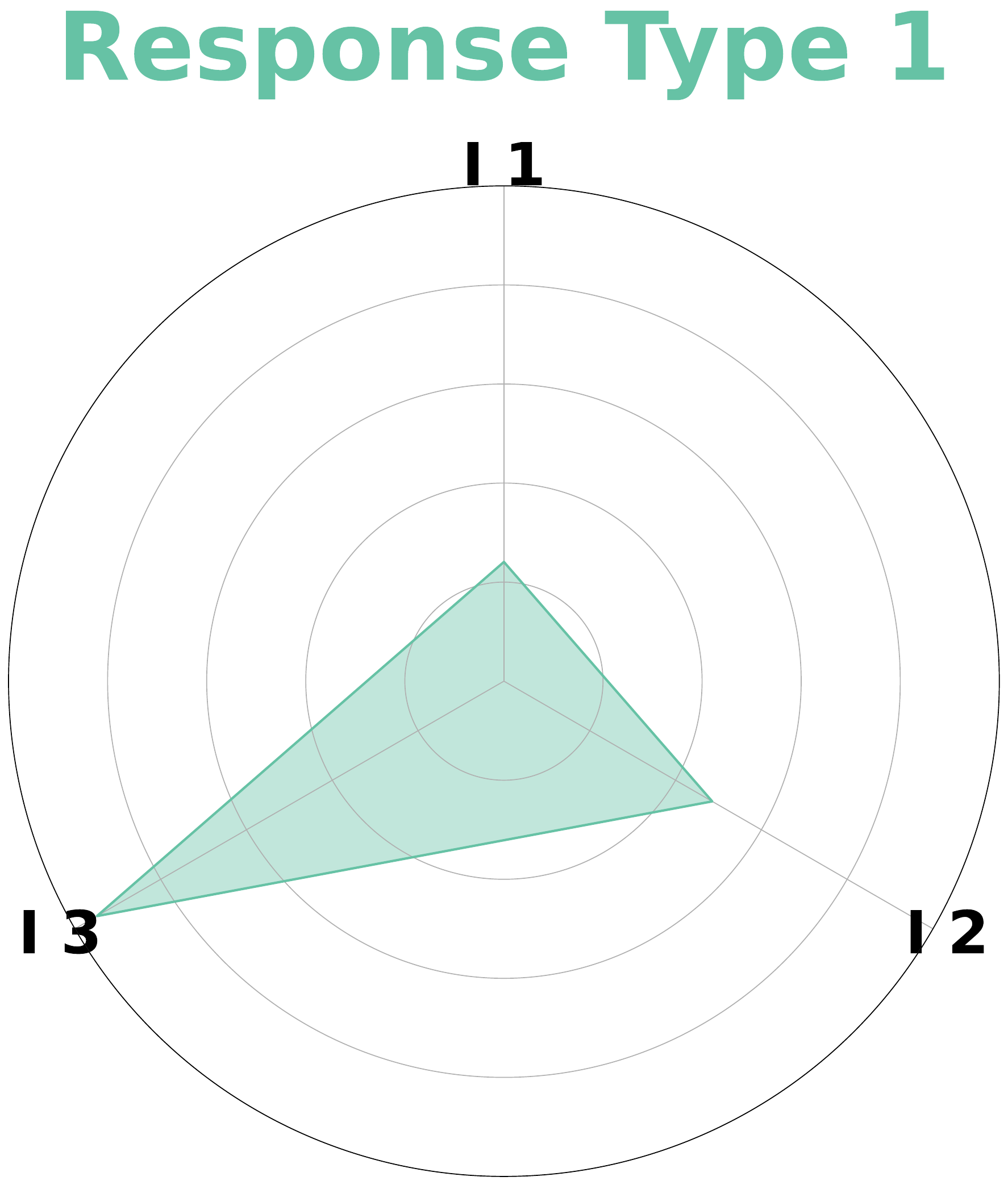}
   \includegraphics[width = 0.19\textwidth]{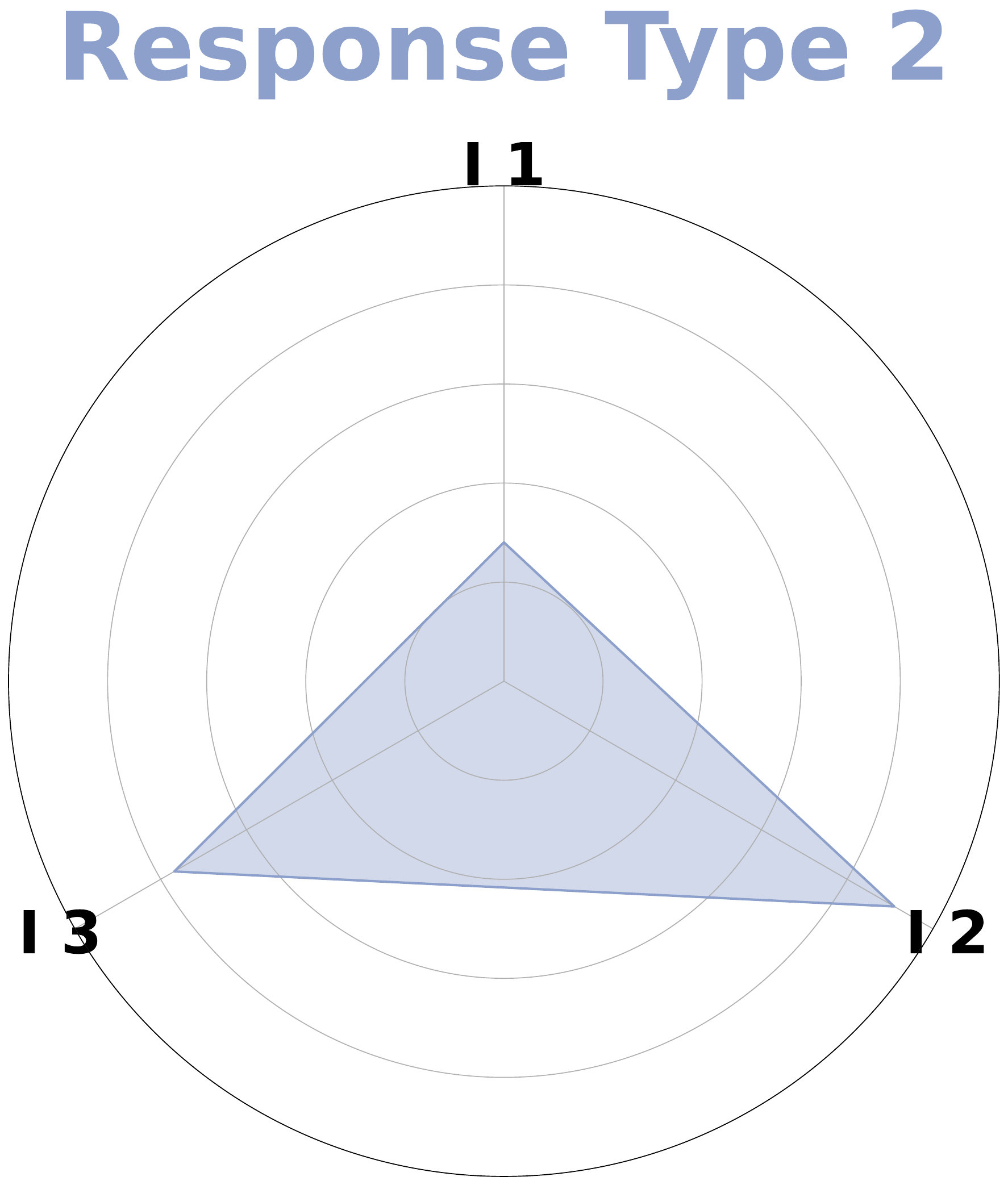}
   \includegraphics[width = 0.19\textwidth]{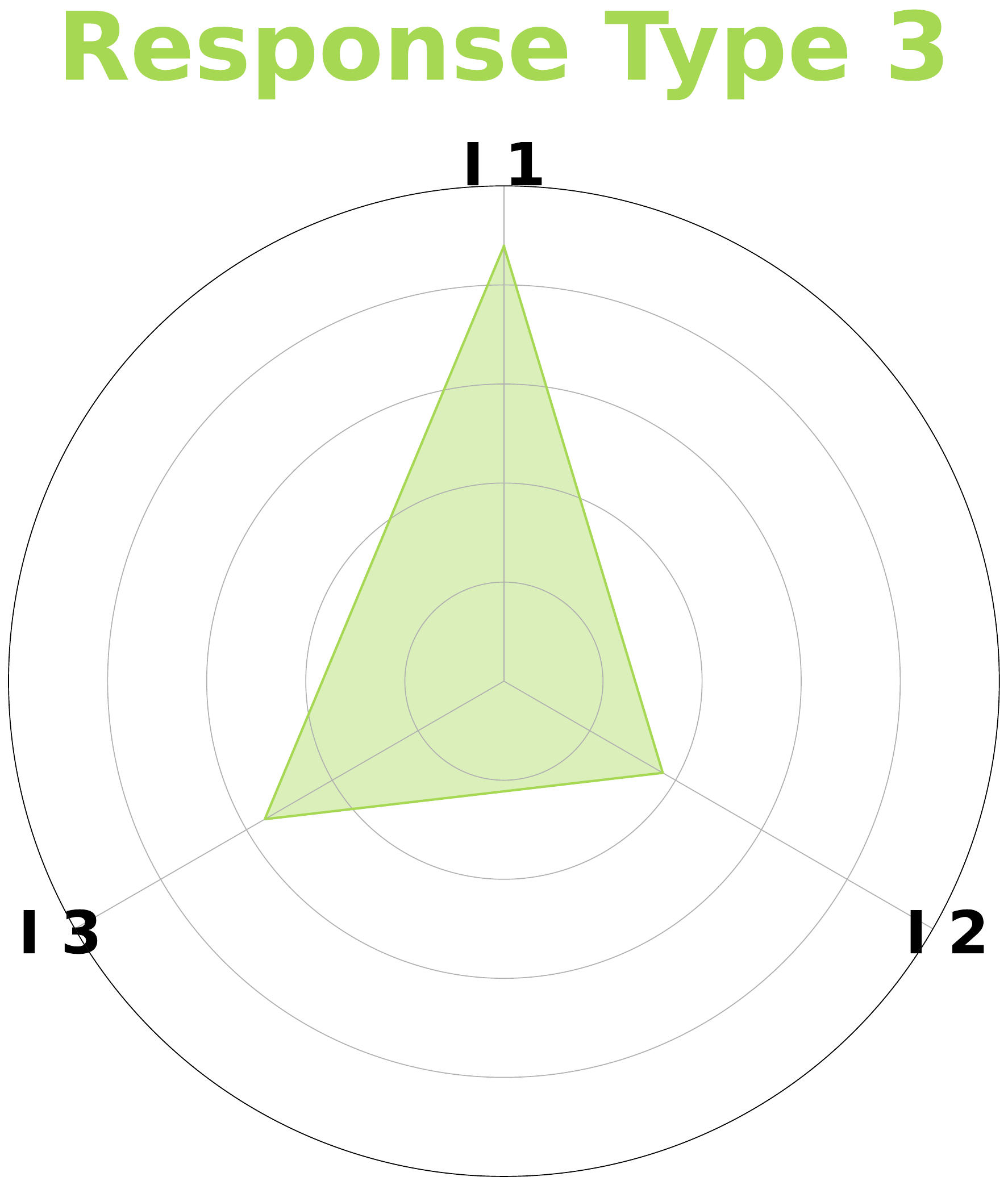}
   \includegraphics[width = 0.19\textwidth]{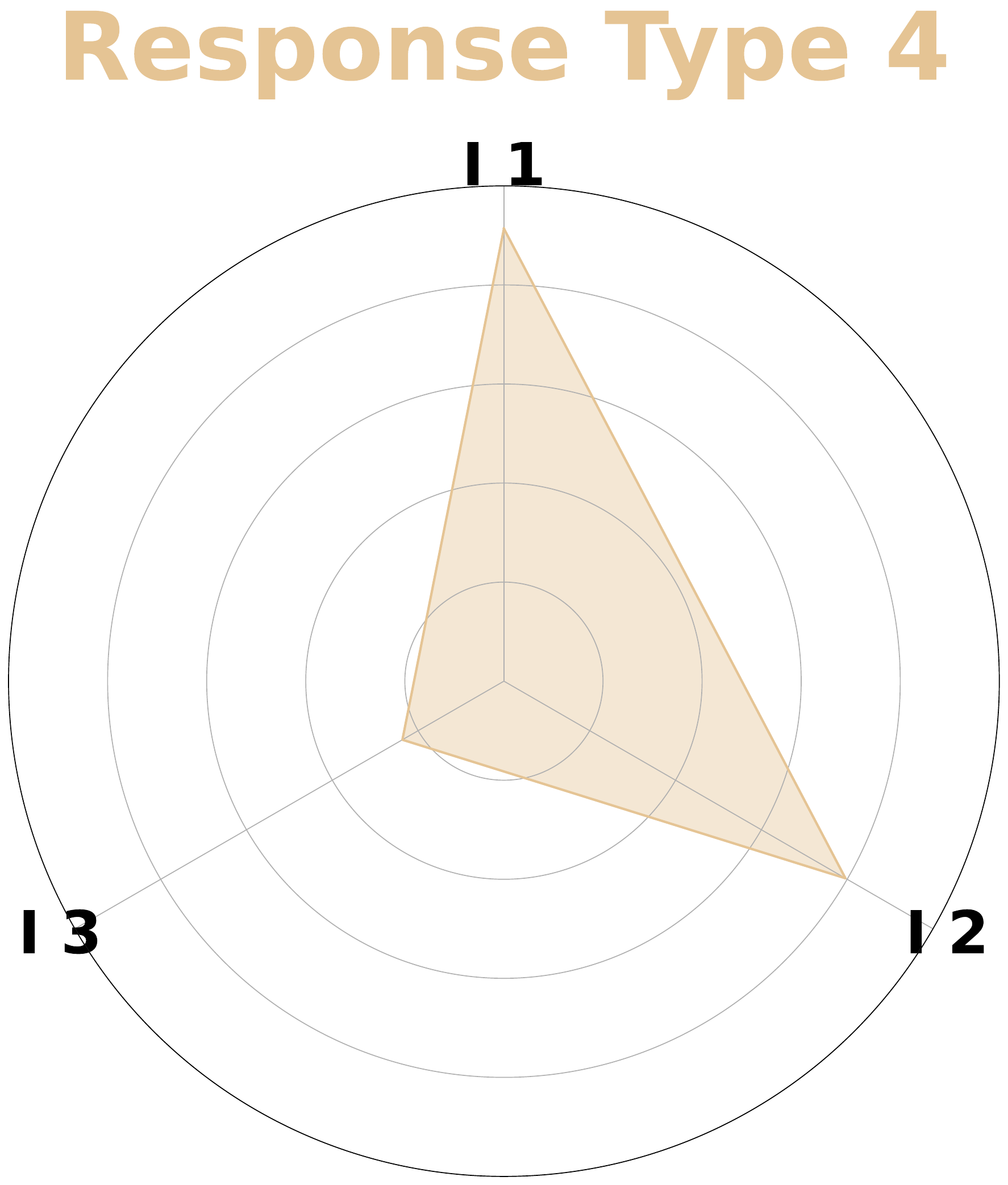}
   \includegraphics[width = 0.19\textwidth]{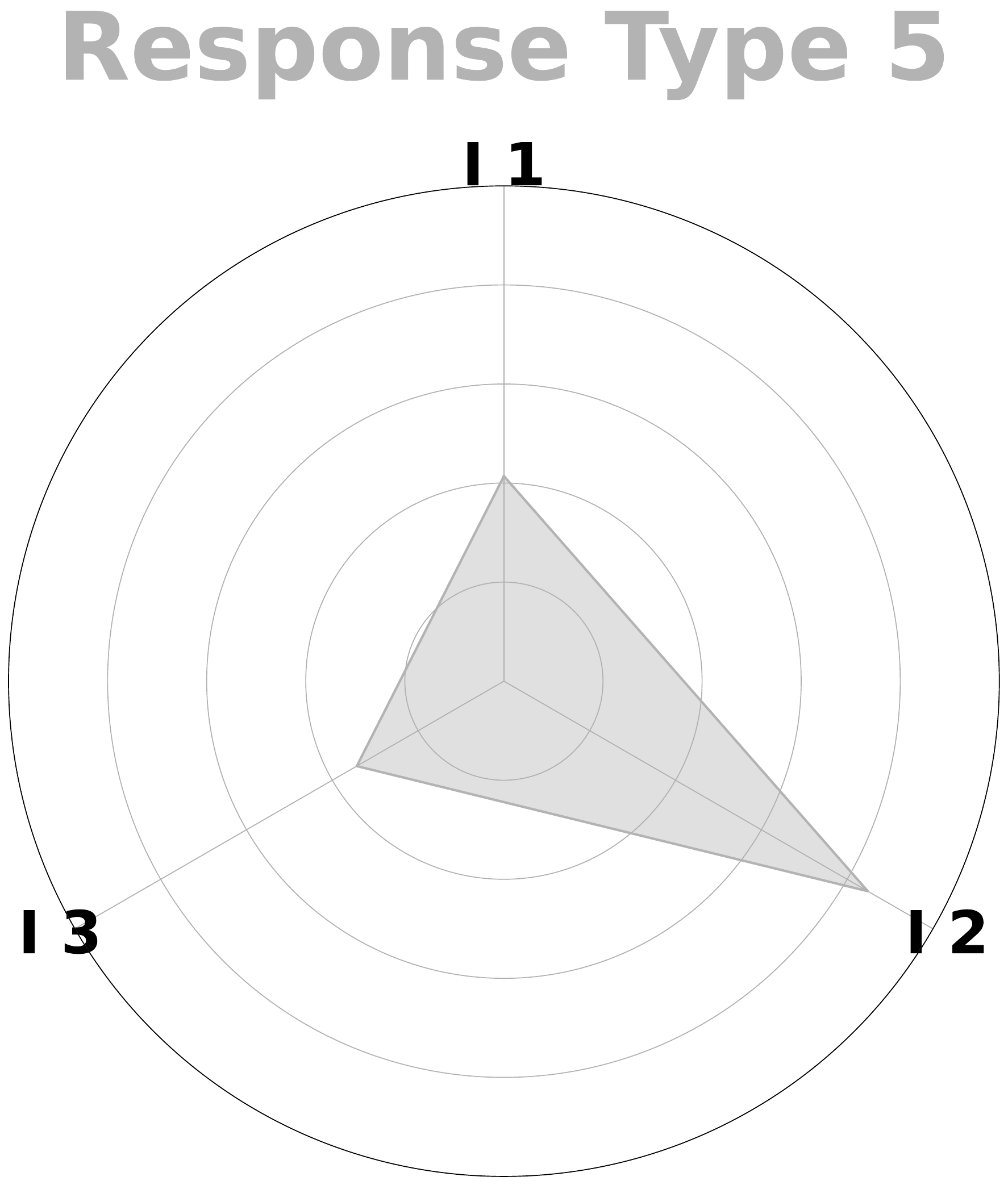}
   \caption{Graphical representation of the gap statistic as well as the dendrogram corresponding to the goodness of the clustering of the questionnaires in data set $\cD_3$. Moreover, the corresponding response types are shown as a spider plot.}
   \label{fig:clusterindices_d3}
\end{figure}

As before, we plot the response types in Figure \ref{fig:clusterindices_d3}. The gap statistic suggests the use of 5 response types, and in the dendrogram, one would choose 5-6 response types. 
Again, this fits well with the actual data generation, which is based on noisy instances of 6 types.
The response types correspond to noisy measurements of five of the six ground-truth values $\sigma_1, \ldots, \sigma_6$, but $\sigma_3 = (3,3,3)$ does not appear as a response type. 
This might well be due to the relatively large noise applied to each coordinate, such that a typical sample from $\sigma_3$ will have different entries.

Regarding the interpretation of the group's fingerprints, we observe that Group 8 contains questionnaires of each response type, Group 9 is mostly concentrated on response types with a high score for item 1, Group 10 is concentrated along those response types in which item 1 is small, and finally, Group 11 has large entries in item 2. 
This reflects the actual data model very well.

\begin{figure}[ht]
   \centering
   \textbf{Group similarity on data set} $\mathbf{\cD_3}$ \par\medskip
   \includegraphics[align=c, width = 0.19\textwidth]{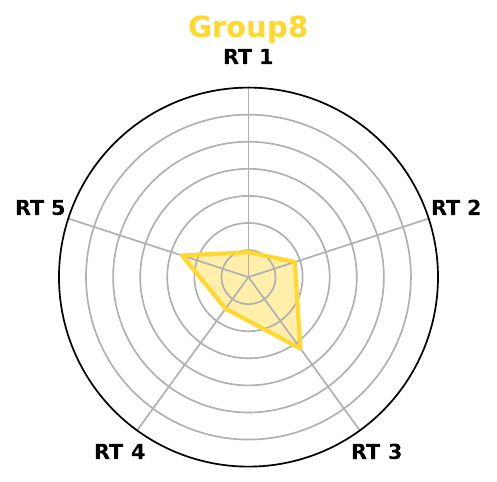}
   \includegraphics[align=c, width = 0.19\textwidth]{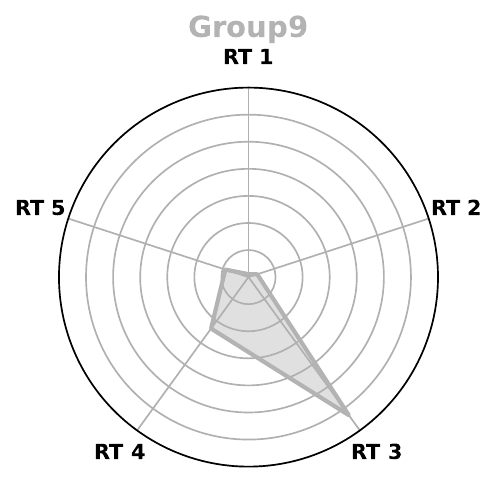}
   \includegraphics[align=c, width = 0.19\textwidth]{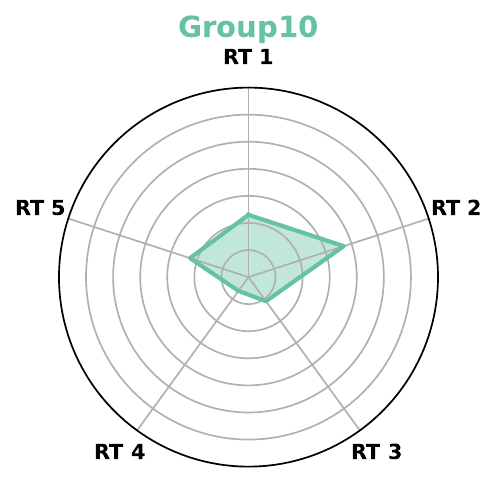}
   \includegraphics[align=c, width = 0.19\textwidth]{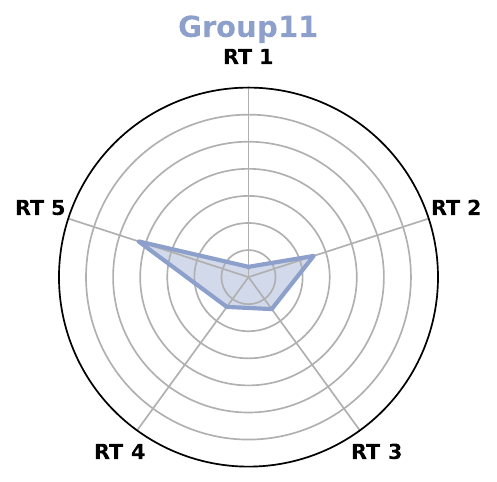}
   \includegraphics[align=c,width = 0.19 \textwidth, height=3cm]{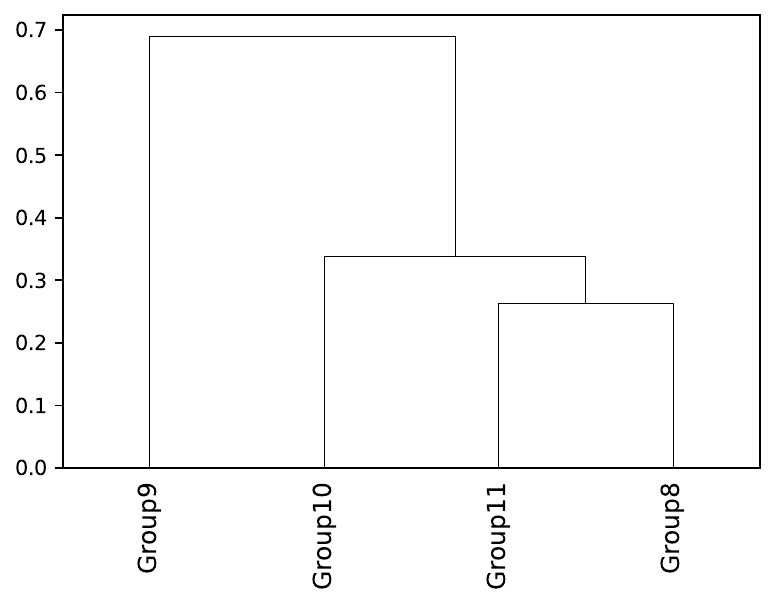}
   \caption{The fingerprints of the different groups regarding the response types as spider plots. The radial $y$-axes are scaled to $(0, 0.7)$. Again, the group similarity on data set $\cD_3$ is visualized by a dendrogram.}
   \label{fig:similarity_d3}
\end{figure}

\subsection{Robustness towards the number of clusters}
\label{sec:more_clusters}
\begin{figure}[ht]
   \centering
   \textbf{Group similarity on data set} $\mathbf{\cD_1}$ \textbf{based on more response types} \par\medskip
   \includegraphics[align=c,width = 0.24 \textwidth]{figures/fingerprints/Fingerprint-Similarity_D1_NCluster_5.pdf}
   \includegraphics[align=c,width = 0.24 \textwidth]{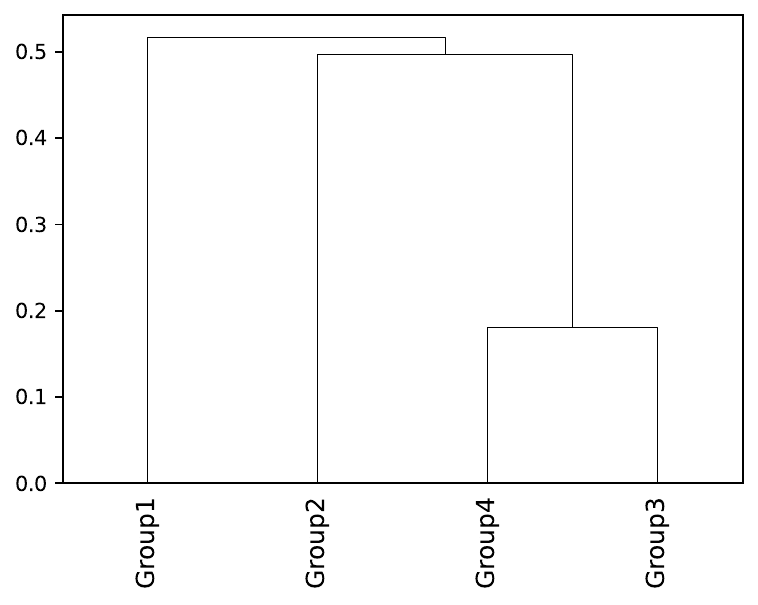}
   \includegraphics[align=c,width = 0.24 \textwidth]{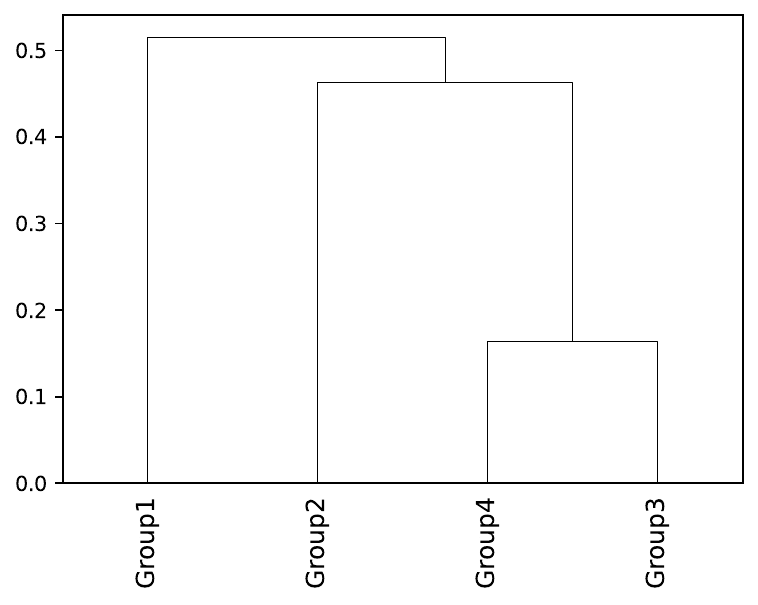}
   \includegraphics[align=c,width = 0.24 \textwidth]{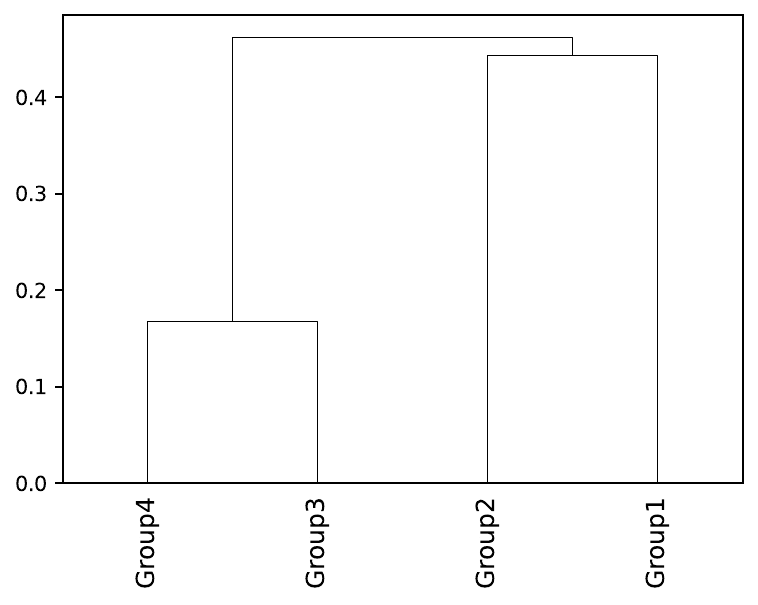}
   \caption{The similarity between the groups in data set $\cD_1$ for a growing number of response types (left: 5 response types, right: 8 response types). The similarity based on the optimal number of response types (5) gives no significantly different result as the similarity based on more response types.}
   \label{fig:similarity_moreclus}
\end{figure}

In this section, we briefly show that a slight over-estimation of the number of clusters does not change the similarity between groups significantly.
In Figure \ref{fig:similarity_moreclus}, we present the dendrogram describing the groups' similarity on data set $\cD_1$ for the optimal choice of 5 response types, as well as for 6, 7, and 8 response types.
As can be easily observed, the similarity between groups does not change significantly.

\begin{figure}[ht]
   \centering
   \textbf{Response types on} $\mathbf{\cD_1}$ \textbf{if more response types are chosen}\par\medskip
   \includegraphics[align=c, width = 0.2\textwidth]{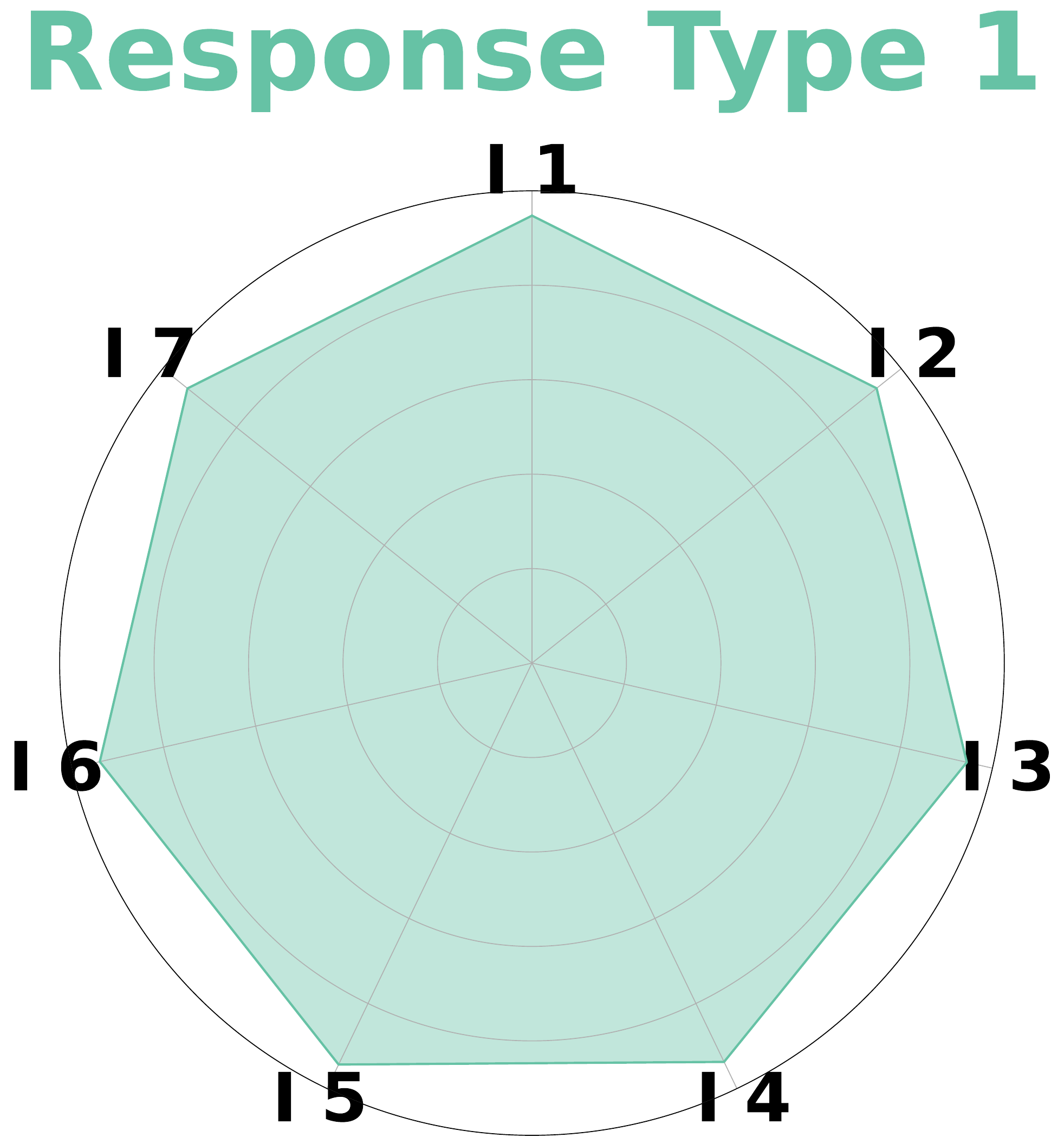}
   \includegraphics[align=c, width = 0.2\textwidth]{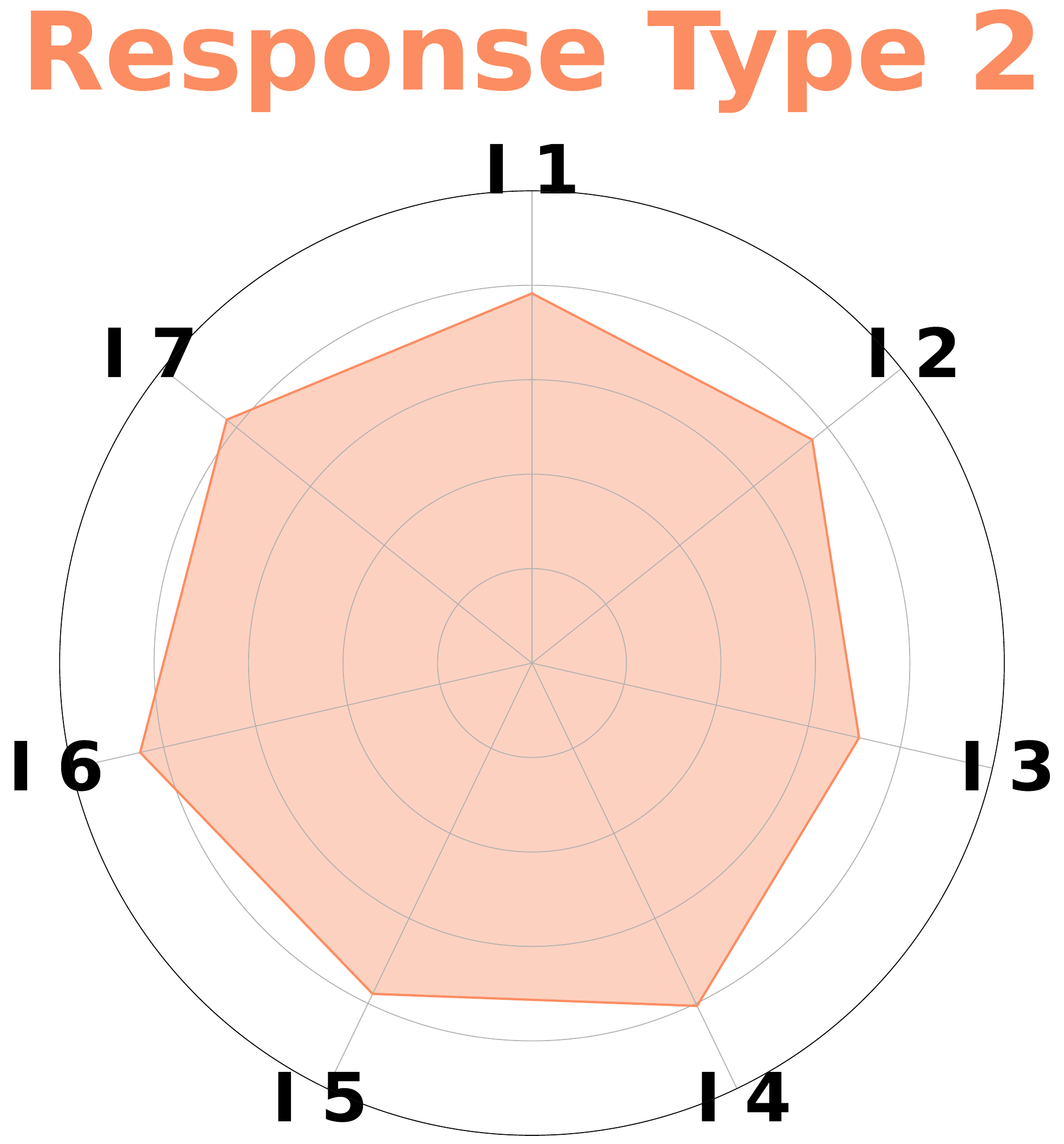}
   \includegraphics[align=c, width = 0.2\textwidth]{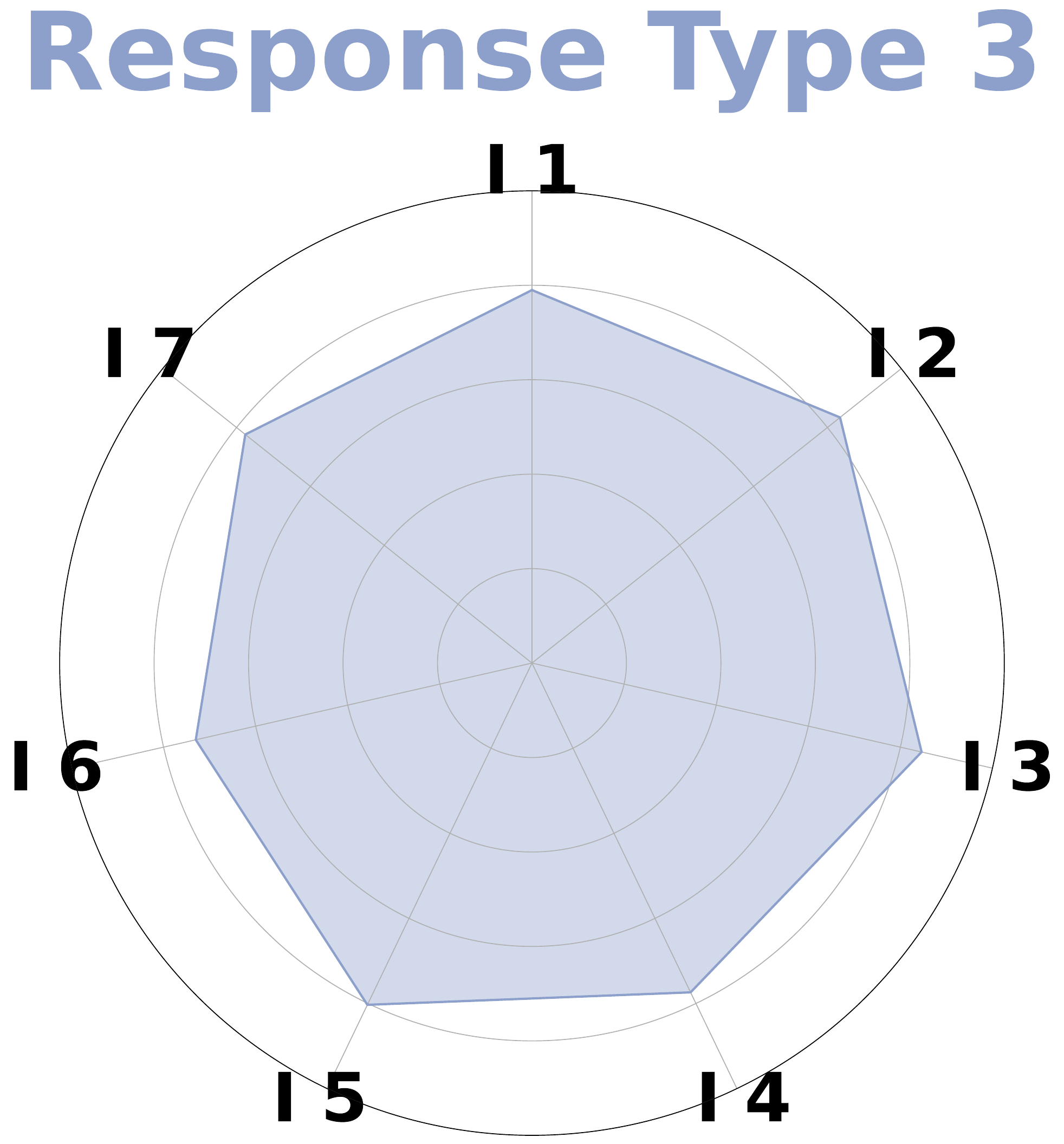}
   \includegraphics[align=c, width = 0.2\textwidth]{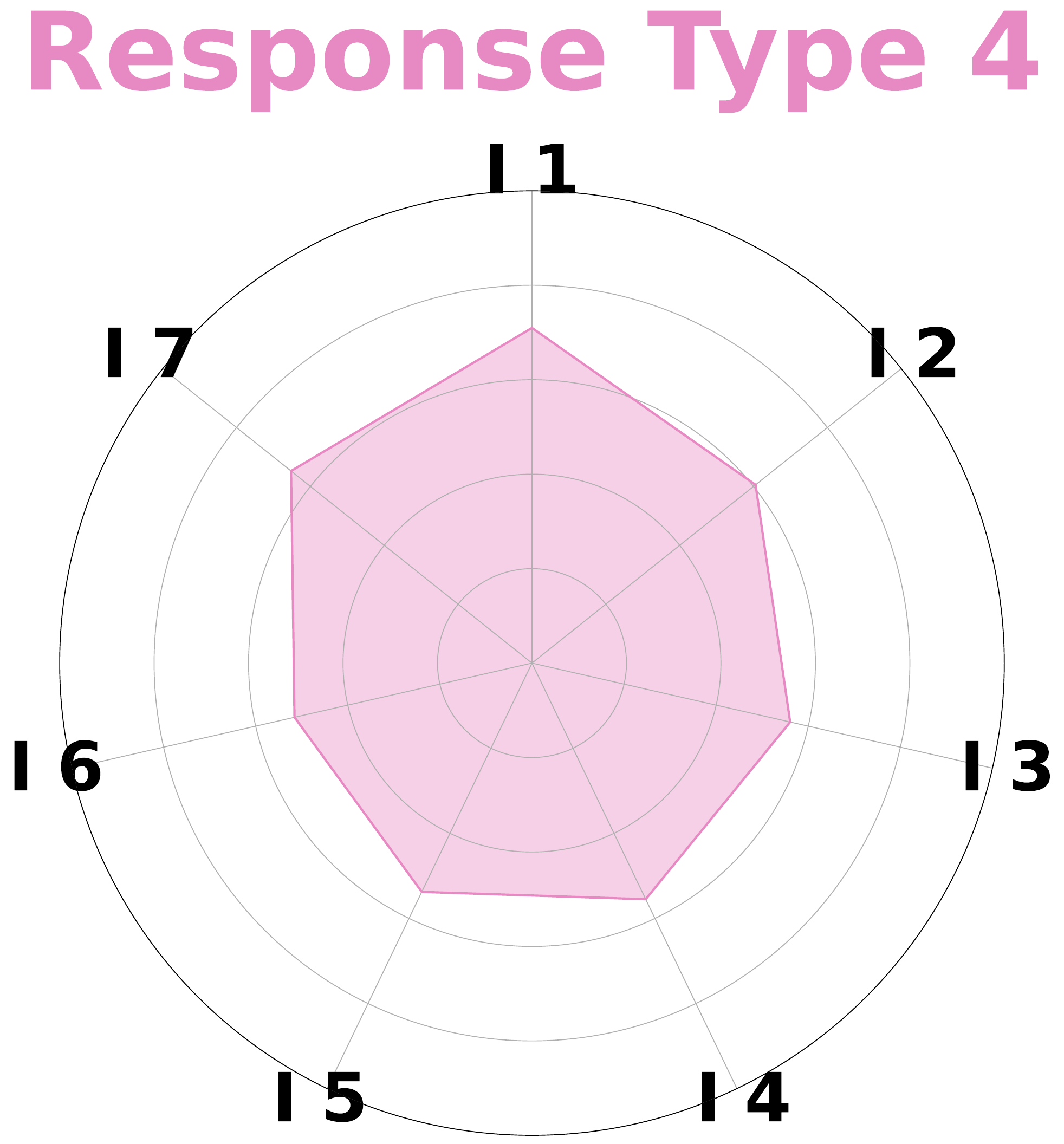}\\
   \vspace*{0.1cm}
   \includegraphics[align=c, width = 0.2\textwidth]{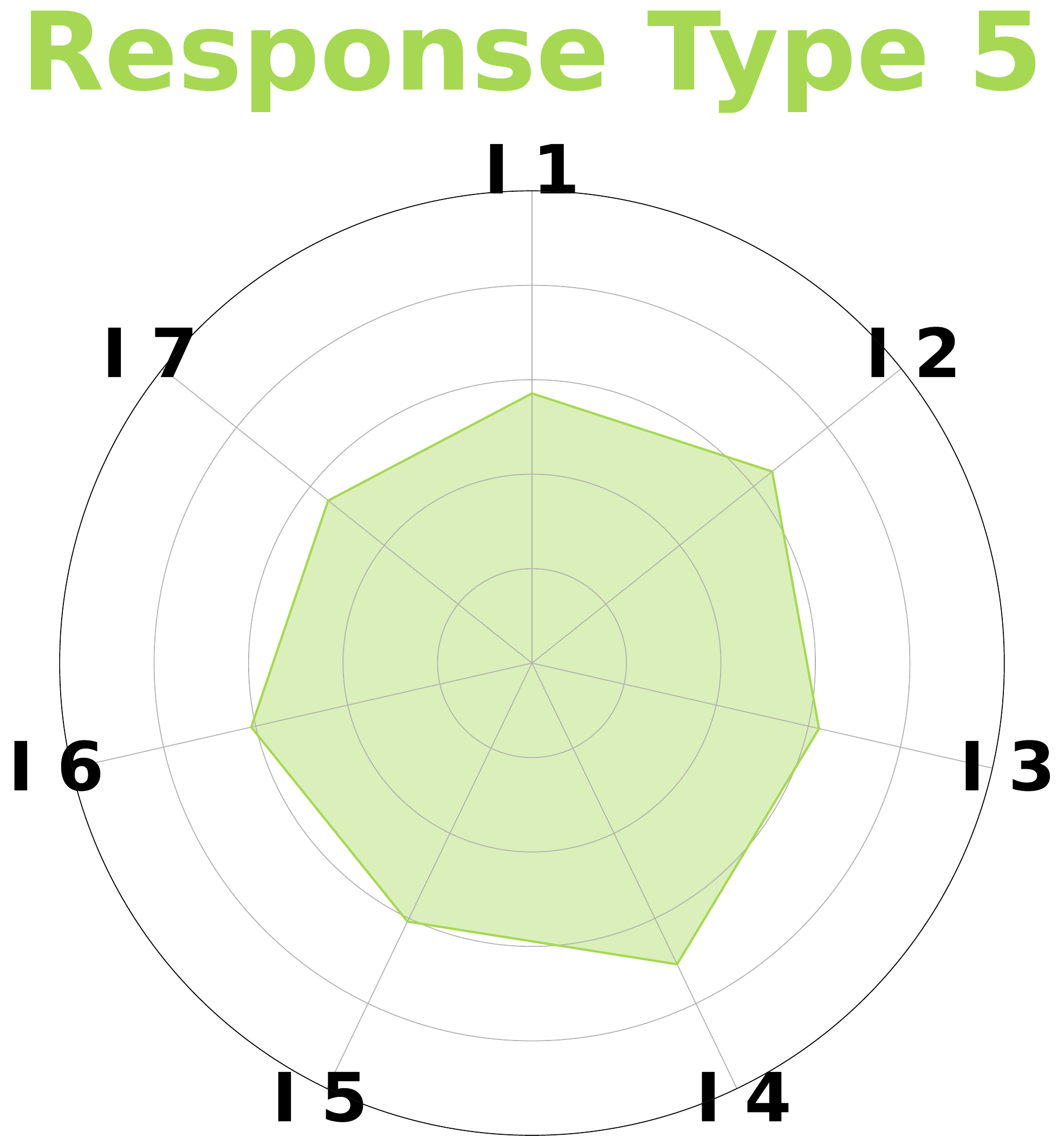}
   \includegraphics[align=c, width = 0.2\textwidth]{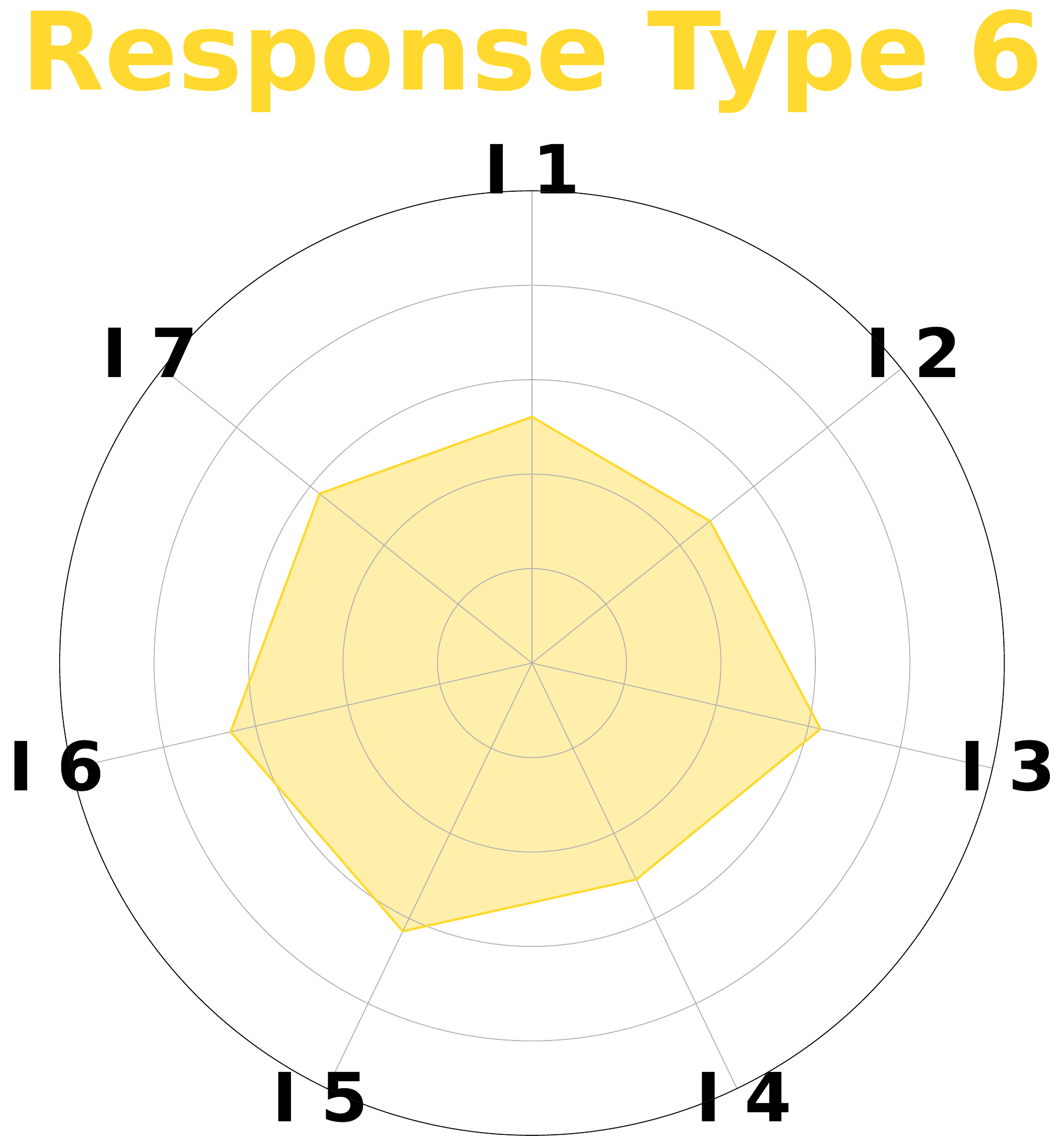}
   \includegraphics[align=c, width = 0.2\textwidth]{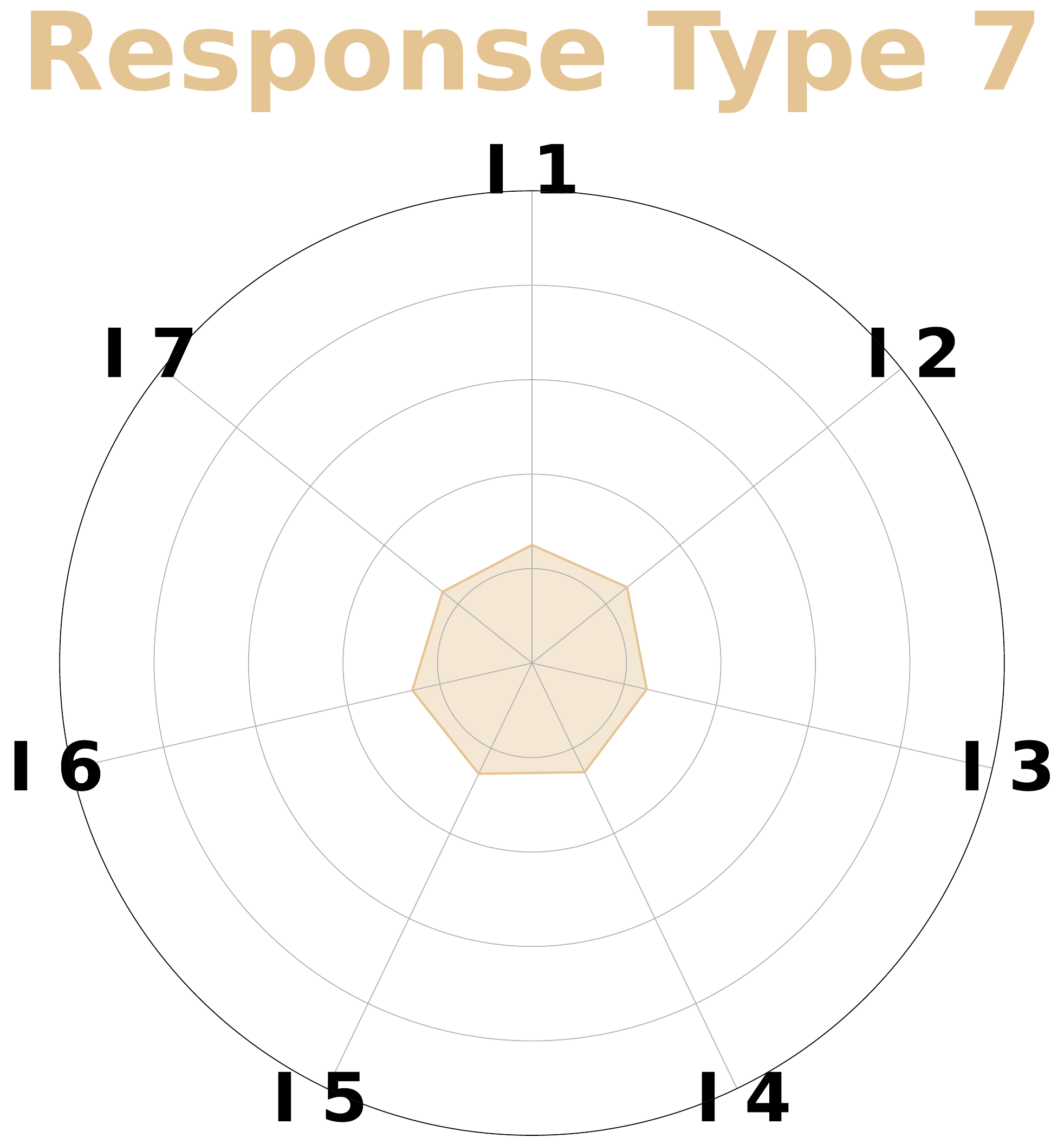}
   \includegraphics[align=c, width = 0.2\textwidth]{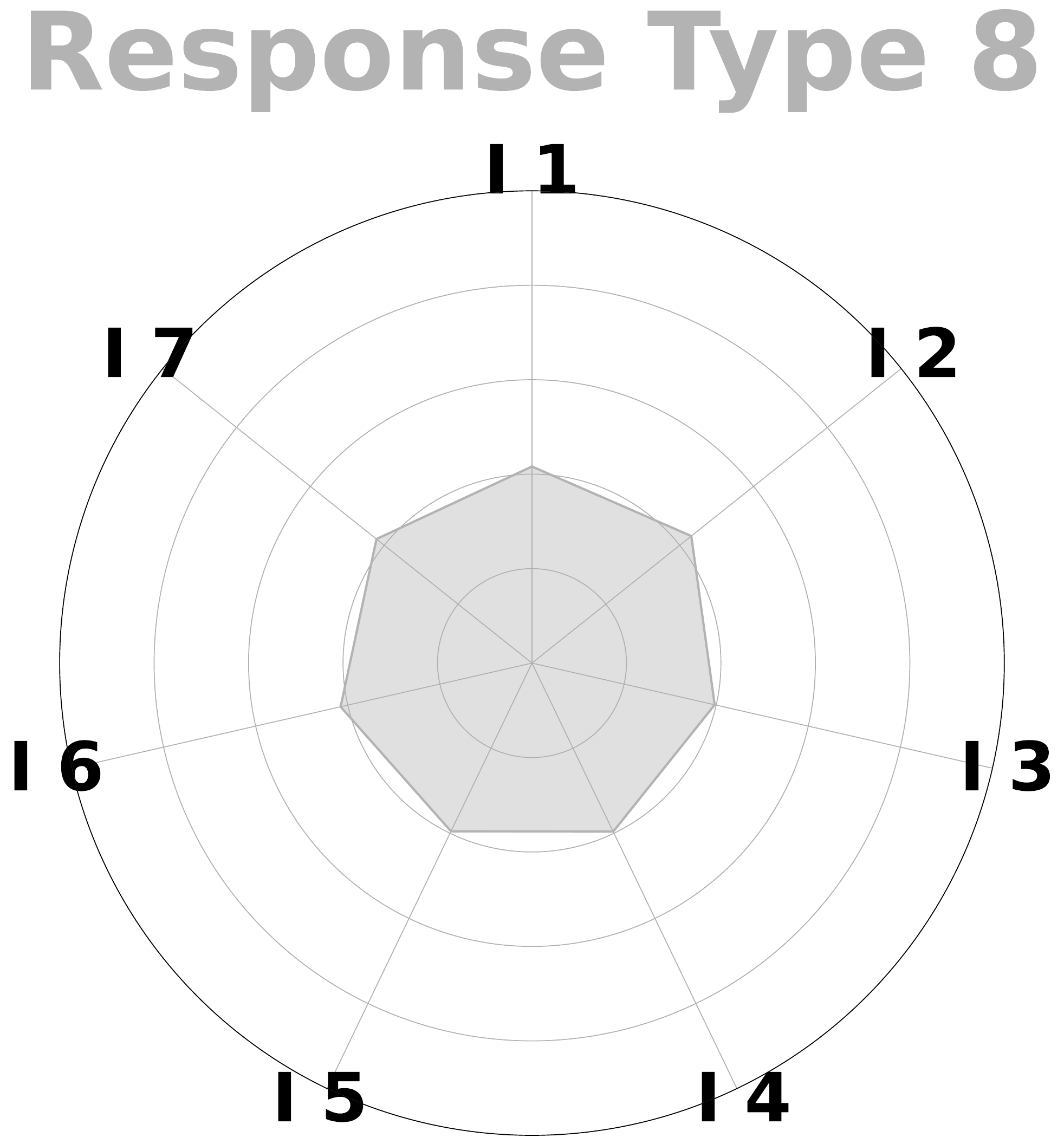}   
   \caption{The eight response types on data set $\cD_1$. While the similarity between group fingerprints does not vary if more response types are used, the response types themselves are much worse separated.}
   \label{fig:responsetypes_moreclus}
\end{figure}

While the similarity itself does not change, it is important to notice that over-estimation of the number of response types has clearly its drawbacks.
The main challenge might arise, as the cluster centroids (the typical questionnaire per cluster), are no longer well separated and potentially harder to explain content-wise.
Recall from Figure \ref{fig:clusterindices_d1} that the five response types were very easy to interpret: they referred to the typical sheets $1...1 - 5...5$ up to some noise.
However, if eight response types are formed, they are not that easy to describe (see Figure \ref{fig:responsetypes_moreclus}).
For example, we observe that response types 4 \& 5 do clearly emerge from the previous response type 3 (see Figure \ref{fig:clusterindices_d1}). 
They contain questionnaires in which the answers are around the typical answer 3, but the cluster centroids are are slightly "deformed" rather than being roughly uniform in all coordinates. 
This is not desirable, because obviously, even if all items measure the same construct and the participant answers with care, the questionnaires $(4,3,3,3,3,2,3)$ (in the cluster of response type 4) and $(3,3,3,4,3,3,3)$ (in the cluster of response type 5) are highly likely to be observed and should, intuitively, both correspond to the same response type.

Tu summarize, it is unproblematic to over-estimate the number of response types with regard to the similarity of the groups, but the response types might become harder to interpret.

\section{Discussion \& Conclusion}
\label{sec:conclusion}

\subsection{On the data preparation step}
\label{sec:conclusion_preparation}
All the steps used in the data preparation step are well known to the data science community, but in the described application, namely the evaluation of questionnaire data, these steps are unlikely to be found.
The first step was to impute missing data by $k$-nearest neighbor imputation.
Normally, such missing data are filled by simply taking the average (either of the row, or of the column) \cite{Troyanskaya2001, Weaver2014}, or simply ignored \cite{Husson2013}, but this does not take into account the dependencies between different items, or reduces the sample size. Therefore, especially since our method is also applicable to questionnaires measuring different constructs with different items, we need a better imputation technique. Since nearest neighbor imputation is well studied and often used for missing data in a variety of data science applications \cite{Troyanskaya2001}, we believe that it should be used in this case as well.

Next, the group samples are balanced by simple oversampling.
This is necessary when the sample sizes of the different groups are different.
For example, suppose one group has a completely different typical response than the other groups, but the number of questionnaires in that group is small.
This tiny fraction of data points won't significantly affect the clustering metrics, and therefore the response type is unlikely to appear as a cluster centroid.
This effect does not occur when the group samples are comparably large.
While oversampling is a standard method in supervised learning, one usually has to be very careful not to overfit a model to a few examples \cite{Chawla_2002, Menardi2012}.
Note that this overfitting effect does not have a serious impact on the proposed method: First, we use the oversampled data set only to identify appropriate response types, while the actual evaluation (e.g., measuring similarity based on the fingerprints) is based only on the actual (not oversampled) data. Second, the obtained clustering is not intended to be applied to unseen data, but only to describe a data set.
We emphasize that an alternative to oversampling would be weighted clustering, but linkage-based clustering algorithms as applied here are known to be incompatible with this approach \cite{Ackerman2021}. However, this choice results in a much higher computational cost, and in applications it may well be that non-linkage-based clustering algorithms could also be used. Furthermore, while oversampling allows for different sample sizes in the different groups, it is very important to note that the sample in each group is representative and valid for the question being studied. 

Finally, the data augmentation step adds some noise to the individual questionnaire items.
Thus, the questionnaire items are no longer integers, but floating-point numbers.
This makes the data matrix more likely to be of full rank, which is crucial for numerical stability, and the resulting clustering more stable against removal of individual data points.
Moreover, it is well known that almost all machine learning models generalize much better when the training data is augmented with random noise \cite{Ding2007, DBLP:conf/icce-tw/ZhangKK20, DBLP:conf/iclr/BelinkovB18, DBLP:conf/acl/MinMDPL20}.
A further idea behind the augmentation is that it turns the naive over-sampling of step two into a SMOTE-similar over-sampling as the single over-sampled data points are subjected to noise and are no simple duplicates of the original data \cite{Chawla_2002}.

\subsection{Comparison to factor analysis and PCA}
\label{sec:conclusion_compare}
First, we observe that certain patterns in the response types correspond to the applicability of the factor analysis approach.
For example, it is highly unlikely that a PCA or factor analysis can be performed meaningfully if the centroids of the clustering of the questionnaire data, i.e. the response types, look \emph{different}. 
In other words, if all response types have a similar shape (as in the set $\cD_1$), or response types 1, 2, and 5 in the data set $\cD_2$, then the principal components between different groups are expected to be the same. 

On data set $\cD_1$, our approach as well as the PCA approach could be safely applied.
It is not very surprising that the hypothesis test following the PCA did not find a significant difference in the group median between Group 3 and Group 4 ($p = 1.00$), if this result is compared to the similarity dendrogram in which those groups are also close to each other. 
However, we observe a striking difference in the "similarity" of Group 2 and Group 3 between the approaches.
Our approach suggests that Group 2 and Group 3 are far apart from each other, while the hypothesis test following the standard PCA approach found no significant differences between these groups ($p = 0.564$).
From a mathematical point of view, this is easy to explain: the hypothesis test is on the \emph{average response} per group (hence a group mean or median, depending on the actual choice of statistical test) and tests the hypothesis that this value is the same between two groups. 
For Group 2 \& 3, this value is expected to be the same (namely 3, based on the data generation). 
However, Group 2 contains mostly moderately answered sheets (expected values between 2 and 4), while Group 3 contains mostly extreme sheets (values either 1-2 or 4-5).
This difference cannot be observed with a mean or median test.
We emphasize that this is not a flaw of either method.
If the hypothesis test result is interpreted correctly, no false conclusions should be drawn.
However, in applications, such a test result is often misleading because it is interpreted as "there are no group differences", which is clearly wrong because it does not take into account the lack of homogeneity of responses in certain groups.
Be that as it may, our approach manages to naturally distinguish between groups with homogeneous and heterogeneous responses, and thus may not lead to false conclusions as easily.

Additionally, our approach gives a very natural and clean way to describe what a \emph{typical answer} is supposed to mean.
When a group's fingerprint is narrowly focused on one response type, that response type is close to the usual group average.
In most cases, however, a fingerprint will have non-vanishing weight on more than one response type. 
In this case, the fingerprint yields several typical response patterns, and the weight indicates what proportion of the questionnaires belong to each typical pattern.
Note that if $f \in [0,1]^\ell$ is the fingerprint of a group and $r_1, \ldots, r_\ell \in \RR^d$ are the response types, the convex combination $\sum f_i r_i$ can be seen as the natural group mean.
Thus, depending on the absolute size of each $f_i$, the fingerprint also directly gives a measure of the variability in the group. For example, if one $f_i = 1$, then only questionnaires of that response type belong to the group, while $f_i \sim \ell^{-1}$ for all $i$ means a uniform distribution across all response types, i.e. a large variation. 
A natural measure of the heterogeneity in a group could therefore be the normalized entropy 
$$H(f) = - \frac{1}{\log \ell} \sum_{i=1}^\ell f_i \log(f_i)$$ of its fingerprint $f$, a quantity that is a very natural measure of variation in a variety of applications from physical systems to information theory \cite{Wehrl1978}, and we propose to use the entropy of fingerprints to measure the heterogeneity of a group. 
A key property of the normalized entropy is that it takes values between $0$ (all questionnaires belong to the same response type)\footnote{As usual, we assume that $0 \log(0) = 0$.} and $1$ (there are equally many questionnaires of each response type), which quantifies homogeneity of a group quite naturally.

\subsection{Advantages and limitations of the proposed approach}
\label{sec:conclusion_limitations}
Probably the most limiting factor of our approach is its purely descriptive nature, i.e. it does not provide a typical measure of significance with respect to group differences. 
In addition, the number of required response types is usually hard to determine.
While the gap statistic yields an algorithmic approach by taking the number of response types at a local maximum in the corresponding scree plot, it is not clear that such a unique or obvious maximum exists. 
Of course, visual inspection of the dendrogram can support the choice, but overall, this freedom of choice might reduce the inter-observer reliability. 

On the positive side, unlike factor analysis or PCA, the proposed method can be applied to any questionnaire data set where differences between multiple groups are to be compared. 
The method comes with a very natural dimensionality reduction from individual questionnaires to fingerprints of groups that have a very natural interpretation, given reasonable response types. 
Moreover, the similarity between groups can be described in a straightforward way. 

\subsection{Overcoming some of the challenges}
\label{sec:conclusion_overcoming}
While the result of the method is indeed descriptive, the obtained similarity between groups can be seen as a quantification of how similar different groups are. When an agglomerative clustering algorithm is applied to the fingerprints, visual inspection of the dendrogram or the formal gap statistic (if the number of groups is large) can be used to identify \emph{clusters} of fingerprints \cite{Tibshirani2001, Mohajer2011}. A difference between two groups exists if and only if these groups do not belong to the same cluster. 
Moreover, as given as an example, the proportion of each response type in a group can be compared to additional describing factors, e.g. country indices via standard methods. As long as the response types can be well interpreted, such analyses can be used to explain the observed effects.

Similarly, while the number of response types used during the algorithm has an impact on the results, we saw in \ref{sec:more_clusters} that the similarity between the fingerprints of groups is robust against choosing too many response types. 
The only effect was observed with respect to the interpretation of the response types (or, the groups' fingerprints on those response types).
The most extreme case is a full clustering, where all possible questionnaire types form a singleton cluster (this corresponds to $n^d$ response types, with $d$ being the number of items and $n$ being the possible number of scores). While the similarity can, in principle, still be measured based on the $n^k$-dimensional fingerprints, an interpretation of the results becomes out of reach.
However, due to the gap statistic backed up by visual inspection of the dendrogram, it is easy to determine a number of response types that is at least close to the potentially optimal choice.
Here, we emphasize that the property of being a \emph{local} maximum of the gap value is important. 
As can be observed in Figure \ref{fig:clusterindices_d3}, a local maximum appears around 5 response types. 
However, for more than 10 response types, the gap value increases again and becomes even larger as the local maximum. As the data set was generated based on noisy instances of 6 base types $\sigma_1, \ldots, \sigma_6$, it is not to be expected that considerably many answer patterns are observed regularly, such that the later response types based on much more then 10 clusters, are likely to be artificial and cannot be well interpreted.

\subsection{Conclusion}
We presented a method to quantify the similarity between different groups based on questionnaire studies, and how it might be possible to explain group differences.
The method does, in contrast to the standard approaches, not require measurement invariance, but it can even use variance in the measurements to better distinguish between the groups.
The approach is easy to apply and relies on very well known data scientific concepts and yields a natural interpretation of the results.
Moreover, we observed that even in situations in which standard factor analyses could be conducted, following simply the standard approach, did not detect all occurring group differences if they are with respect to group homogeneity or heterogeneity rather than based on the average answer. 
Overall, we believe that the proposed approach may help a variety of applicants to analyze their complex data sets.

\section*{Funding Information}
We gained support from the von Opel Hessische Zoostiftung and the OA Open Access publication fund of Goethe University Frankfurt.

\section*{Competing Interests}
The authors have no competing interests to declare.

\printbibliography

\end{document}